\documentclass[lettersize,journal]{IEEEtran}
\usepackage{amsmath,amsfonts}
\usepackage{algorithmic}
\usepackage{array}
\usepackage[caption=false,font=normalsize,labelfont=sf,textfont=sf]{subfig}
\usepackage{textcomp}
\usepackage{stfloats}
\usepackage{url}
\usepackage{enumitem}
\usepackage{verbatim}
\usepackage{wasysym} 
\usepackage{graphicx}

\usepackage{bbding} 
\usepackage{booktabs}
\usepackage{makecell}
\usepackage{multirow}
\usepackage[linesnumbered,ruled]{algorithm2e}
\def\BibTeX{{\rm B\kern-.05em{\sc i\kern-.025em b}\kern-.08em
    T\kern-.1667em\lower.7ex\hbox{E}\kern-.125emX}}
\usepackage{balance}
\usepackage{orcidlink}
\begin{document}
\title{MKF-ADS: Multi-Knowledge Fusion Based Self-supervised Anomaly Detection System for Control Area Network}
\author{Pengzhou Cheng~\orcidlink{0000-0003-0945-8790}, Zongru Wu, and Gongshen Liu
\thanks{This research was supported by the following funds: National Key R\&D Program of China (Grant No. 2023YF3303805), the Joint Funds of the National Natural Science Foundation of China (Grant No. U21B2020). (Corresponding author: lgshen@sjtu.edu.cn)}

\thanks{Pengzhou Cheng, Zongru Wu, and Gongshen Liu are with the Department of Electronic Information and
Electrical Engineering, Shanghai Jiao Tong University, Shanghai, 201100,
China (e-mail: pengzhouchengai@gmail.com, lgshen@sjtu.edu.cn).}

}
\newcommand{\MySquare}{\@chooseSymbol{'146}}
\markboth{IEEE TRANSACTIONS ON VEHICULAR TECHNOLOGY}%
{How to Use the IEEEtran \LaTeX \ Templates}

\maketitle

\begin{abstract}
Control Area Network (CAN) is an essential communication protocol that interacts between Electronic Control Units (ECUs) in the vehicular network. However, CAN is facing stringent security challenges due to innate security risks. Intrusion detection systems (IDSs) are a crucial safety component in remediating Vehicular Electronics and Systems vulnerabilities. However, existing IDSs fail to identify complexity attacks and have higher false alarms owing to capability bottleneck. In this paper, we propose a self-supervised multi-knowledge fused anomaly detection model, called MKF-ADS. Specifically, the method designs an integration framework, including spatial-temporal correlation with an attention mechanism (STcAM) module and patch sparse-transformer module (PatchST). The STcAM with fine-pruning uses one-dimensional convolution (Conv1D) to extract spatial features and subsequently utilizes the Bidirectional Long Short Term Memory (Bi-LSTM) to extract the temporal features, where the attention mechanism will focus on the important time steps. Meanwhile, the PatchST captures the combined contextual features from independent univariate time series. Finally, the proposed method is based on knowledge distillation to STcAM as a student model for learning intrinsic knowledge and cross the ability to mimic PatchST. We conduct extensive experiments on six simulation attack scenarios across various CAN IDs and time steps, and two real attack scenarios, which present a competitive prediction and detection performance. Compared with the baseline in the same paradigm, the error rate and FAR are  2.62\% and 2.41\% and achieve a promising F1-score of 97.3\%.
\end{abstract}


\section{Introduction}
\IEEEPARstart{C}{ontrol} Area Network (CAN) has become the standard communication protocol for vehicle networks, playing a decision-making role in the interaction of multiple Electronic Control Units (ECUs)~\cite{9320546}. The emergence of Intelligent Connected Vehicles (ICVs) has led researchers to focus more on CAN. Therein, interconnectivity requires CAN to open additional interfaces to achieve intelligent services, such as autonomous vehicles (AVs), vehicular network communication, and cloud computing~\cite{agrawal2022novelads, han2021ppm, 8514157}. However, these intelligent services also make the otherwise closed communication in-vehicle networks (IVNs) confront attacks from the inter-extra~\cite{liu2021blockchain}. Unfortunately, the CAN bus was not designed with any security mechanisms in mind (e.g., authentication and encryption)~\cite{pawelec2019towards}. When communicating via the bus or with external devices via Bluetooth or hardware ports~\cite{yang2021mth, zhang2021many,boualouache2023survey}, the bus or ECUs may be hijacked, which will pose serious repercussions to the lives of drivers and pedestrians~\cite{ashraf2020novel, wang2023periodic, park2022unsupervised}. Therefore, vehicular network security is a crucial research area. 

Although taking purpose-direction to security problems is the most straightforward (e.g., encryption or authentication)~\cite{luo2023esia}, more work has proven that such approaches consume a lot of memory and computation~\cite{schell2020valid, foruhandeh2019simple, kneib2020easi}. Moreover, such approaches require consideration of backward compatibility concerns~\cite{cheng2023desc}. In contrast, the in-vehicle intrusion detection systems (IV-IDS) are an efficient protection mechanism used to detect cyber-attacks on automobiles~\cite{aliwa2021cyberattacks}, which has two main categories, signature-based and anomaly-based~\cite{erlacher2020high, wu2022rtids}. Signature-based IDS use supervised learning to capture fixed types of attacks from existing databases. However, the method requires a large amount of marker data, while real-life CAN attack data is not readily available. As anomaly-based IDS can detect unknown malicious network activities, especially zero-day attacks, it is more suitable for the open and complex vehicular network environment~\cite{liu2022response}. 

Some ADS research has begun to model the inherent correlation of the normal CAN messages, so as to identify a deviation in detecting malicious messages. For instance, Song~\textit{et al.}~\cite{song2021self} presented an anomaly method that identifies attack messages through noisy pseudo-normal data. However, this method is supervised and thus requires abnormal datasets in the learning process. In contrast, Talyor~\textit{et al}~\cite{taylor2016anomaly} and Qin~\textit{et al.} are both used for long and short-term memory (LSTM) to learn the temporal correlation of CAN transition bits, in order to search for a deviation between abnormal and normal messages. But they made the model at all bits (usually 64 bits) without considering the actual meaning, resulting in the predicted performance being relatively worse. Meanwhile, the Convolutional Neural Network (CNN) and attention mechanism have been introduced as a state-of-the-art paradigm aimed at capturing comprehensive features~\cite{hanselmann2020canet, sun2021anomaly}. However, they fail to capture long time-series historical dependencies. Recently, Alkhatib \textit{et al.}~\cite{alkhatib2022can} designed a CAN-BERT model that learns the context semantic features of CAN ID sequences. Nam \textit{et al.}~\cite{nam2021intrusion} regarded the CAN ID sequence as a sentence with several words and adopted the Generative Pretrained Transformer (GPT) model to learn the normal pattern of each sequence. Although contextual features are important for ADS, these advanced transformer-based models cannot satisfy IVNs with limited resources due to the large number of parameters. 

To address the above challenges, we propose a \textbf{M}ulti-\textbf{K}nowledge \textbf{F}usion based self-supervised \textbf{A}nomaly \textbf{D}etection \textbf{S}ystem, called MKF-ADS. The method is self-supervised to detect malicious attacks from deviations learned from normal messages, which follows the paradigm of multivariate time series prediction. The deviation is built on the actual signals and prediction signals.     In order to construct a more realistic and meaningful model, the CAN payload is preprocessed using the "READ" method, thereby determining the signal boundaries and extracting the continuous physical signal.  The decision reference of MKF-ADS is from multi-knowledge, consisting of spatial-temporal correlation and contextual information. The fusion mechanism is a guidance and imitation mode from high to low due to the resource limit of IVN. Extensive experiments have shown that the proposed model can reduce the error rate and FAR to 2.62\% and 2.41\%, respectively. Compared with the baseline as the same paradigm, our method achieves a competitive F1-score of 97.3\%. The main contributions of this paper are as follows.
\begin{enumerate}
    \item We propose MKF-ADS, a self-supervised anomaly prediction model by fusing multi-knowledge for CAN. The framework consists of the STcAM and PatchST modules, used for extracting spatial-temporal correlation knowledge and long time-series historical dependency knowledge.
    \item We design a lightweight STcAM module, where the Conv1D component first extracts spatial features, BiLSTM subsequently captures temporal features, and the soft attention mechanism focuses on the important times step.  Moreover, the PatchST is introduced to separate multivariate time series and captures context features on all-time steps.
    \item We construct a multi-knowledge fusion framework, which regards the StcAM as a student model and the PatchST as a teacher model.   STcAM not only learns inherent knowledge but also cross-mimicks valuable features from PathchST, which avoids conflict on different knowledge and improves the training stability.
    \item We evaluate the prediction and detection performance of six intrusion scenarios and then generalization performance is verified on different CAN IDs, multi-time steps, and real-attack tasks in ROAD tasks. Also, The feasibility of the model's complexity and overall efficiency is discussed when applied in the vehicular network.
\end{enumerate}

The rest of this paper is organized as follows. Section ~\ref{sec2} reviews the research progress and discusses these with our work in depth. In Section ~\ref{sec3}, the preparatory knowledge about IVNs, attack models, and design intuition is introduced. In Section ~\ref{sec4}, we present the MKF-ADS details. In Section ~\ref{sec5}, we present the experimental results along with a discussion of the performance and efficiency advantages. Section ~\ref{sec6} concludes our work.

\section{Related Work}\label{sec2}
\begin{table*}[t]
\caption{COMPARISON OF METHODS WITH OUR PROPOSED MKF-ADS AND EXISTING SCHEMES (SA: STATISTICAL ANALYSIS, ML: MACHINE LEARNING, DL: DEEP LEARNING)}
\centering
\Large
\label{tab1}
\renewcommand\arraystretch{1.5}
\resizebox{\linewidth}{!}{
\begin{tabular}{l|c|c|c|c|c|c|c}
\hline
Methods & \makecell[c]{\cite{muter2011entropy, song2016intrusion} \\ \cite{studnia2018language, olufowobi2019saiducant}}  & \cite{tanksale2019intrusion,dey2023efficient} & \makecell[c]{\cite{song2020vehicle, xun2021deep, lo2022hybrid} \\ \cite{cheng2022stc, javed2021canintelliids}} & \makecell[c]{\cite{agrawal2022novelads,jeong2023x,tariq2020cantransfer} \\ \cite{ashraf2020novel,zhang2023federated}} & \cite{sun2021anomaly, taylor2016anomaly, qin2021application, zhang2019deep} & \cite{alkhatib2022can, nam2021intrusion} & MKF-ADS (Ours)\\ \hline
Categories & SL & ML & \makecell[c]{DL-based \\supervised-learning} & \makecell[c]{DL-based anomaly\\ classification} & \makecell[c]{DL-based\\ anomaly prediction} & \makecell[c]{Transformer-based \\ DL} & \makecell[c]{DL-based\\ anomaly prediction} \\ \hline
Technology & \makecell[c]{Entropy \\and frequency} & \makecell[c]{HMM, SVM,\\and XGBoost}  & \makecell[c]{CNN, LSTM, GRU, \\and AM} & \makecell[c]{CNN, LSTM, \\and One-class} & \makecell[c]{CNN, LSTM, \\and AM} & BERT and GPT & \makecell[c]{Multi-knowledge \\ fused} \\ \hline
Attack threats & Low & Low & Medium & Medium & High & Medium & High \\ \hline
Knowledge modeling & \XSolidBrush & \XSolidBrush & \Checkmark\kern-1.2ex\raisebox{1ex}{\rotatebox[origin=c]{125}{\textbf{--}}} &\Checkmark\kern-1.2ex\raisebox{1ex}{\rotatebox[origin=c]{125}{\textbf{--}}} & \Checkmark\kern-1.2ex\raisebox{1ex}{\rotatebox[origin=c]{125}{\textbf{--}}} & \XSolidBrush & \CheckmarkBold\\ \hline
Computation complexity & Low & Low & High & Medium & Medium & High & Low \\ \hline
Collection delay & Low & Low & High & High & Low & High & Low \\ \hline
Detection Capability (1-5) & \FiveStar & \FiveStar\FiveStar & \FiveStar\FiveStar\FiveStar &\FiveStar\FiveStar\FiveStar & \FiveStar\FiveStar\FiveStar & \FiveStar\FiveStar\FiveStar\FiveStar & \FiveStar\FiveStar\FiveStar\FiveStar\FiveStar \\
\hline
\end{tabular}}
\end{table*}
This section provides an in-depth overview of previous works on vehicular network security, broadly categorized into authentication-based, cryptography-based, and IDS-based. 

\subsection{Security Models for Vehicular Network}
Authentication-based and cryptography-based methods have been studied over the years by several researchers~\cite{li2022sustainable, shawky2023efficient}.  dariz~\textit{et al.}~\cite{dariz2017trade} provide a combined encryption and authentication mechanisms for vehicular network, and in~\cite{farag2017cantrack} uses a dynamic symmetric key to encrypt the 8byte data payload but does not modify the CAN ID. Although these methods provide a certain security guarantee for IVN, they also suffer from response delays or increased bus loads~\cite{bella2019toucan}. Alladi~\textit{et al.}~\cite{alladi2020lightweight} proposed a lightweight and secure authentication and attestation scheme for attesting vehicles while they are on the roads. Cui~\textit{et al.}~\cite{cui2023lightweight} proposed a lightweight encryption and authentication scheme for the CAN bus of AVs using message authentication codes and Grain stream cipher. In short,  they are either computationally demanding or protocol-modified, which prevents them from providing dependable efficient defense. Hence, providing security defense via MACs or encryption is not a straightforward proposition. 

Intrusion Detection Systems (IDSs) have been widely studied owing to taking positive countermeasures in the event of attacks~\cite{xun2023side}. Early IVN-IDSs identify malicious messages through a predefined knowledge base (e.g., frequency and entropy)~\cite{song2016intrusion,studnia2018language, olufowobi2019saiducant}, but these methods have a high false positive rate (FPR) and cannot detect unknown attacks. Generally, statistics-based IDS are efficient and lightweight but may ignore low-volume and aperiodic frequency attacks. Machine learning-based methods have been extensively researched~\cite{tanksale2019intrusion, dey2023efficient}. However, these approaches have a botnack in detection performance. Due to the great density of the generated vehicular data and enhanced computing resources, various deep learning-based strategies have recently been presented that outperform the current signature-based or machine-learning approaches. Song \textit{et al}.~\cite{song2020vehicle} designed a deep convolution neural network (DCNN) based method for anomaly detection, removing the need for extracting handcrafted features. To extract valuable features, the CNN and LSTM-based deep learning model was presented by Xun \textit{et al}.~\cite{xun2021deep}. The hybrid model was also introduced in ~\cite{lo2022hybrid}. Further, Cheng \textit{et al}.~\cite{cheng2022stc} incorporated attention mechanisms on this basis to reduce the false alarm rate. Javed \textit{et al}.~\cite{javed2021canintelliids} replaced the time-permissive feature extraction model LSTM with the GRU model to improve the inference speed. Although the detection capabilities of these methods have improved to varying degrees, they all rely heavily on a large number of pre-labeled labels for supervised learning and are not practical in defending against attack variants. 

Anomaly detection systems (ADSs) have been struggling to break through on how to model unknown attacks. Ashraf \textit{et al}.~\cite{ashraf2020novel} utilized auto-encoder (AE) to reconstruct the input messages via spatial-temporal features. Similarly, Agrawal \textit{et al}.~\cite{agrawal2022novelads} presented a method that reconstructs sequence directly using spatial-temporal features and then detects anomalies based on threshold deviations. Jeong~\textit{et al.}~\cite{jeong2023x} present explainable intrusion detection based on the self-supervised AE paradigm. Tariq \textit{et al}.~\cite{tariq2020cantransfer} trained a Convolutional LSTM-based model by transfer learning to identify attacks, called CANTrasfer. They also applied one-shot learning so the system could detect adaptive attacks, but the detection performance is relatively poor. Zhang~\textit{et al.}~\cite{zhang2023federated} presented graph convolutional neural network (GCNN)-based ADS to capture the falsification attacks. Considering that CAN messages always follow comparatively fixed patterns, ADS can use predictive analysis to detect complex cyber attacks. This paradigm is stringent excluding all suspicious messages that deviate from the prediction. Taylor~\textit{et al.}~\cite{taylor2016anomaly} and Qin~\textit{et al.}~\cite{qin2021application} developed the LSTM-based model to predict the next package, where the error measure is made using the predicted output value of the log loss function. Sun~\textit{et al.}~\cite{sun2021anomaly} introduced convolution and attention mechanism (AM) to make up for valuable feature extraction. Interestingly, In ~\cite{alkhatib2022can} and ~\cite{nam2021intrusion}, the outstanding model in NLP has been suggested applying to ADS (e.g. BERT~\cite{alkhatib2022can} and GPT~\cite{nam2021intrusion}). Although the plentiful context features are incorporated, the number of model parameters is tens to hundreds of times higher than the previous scheme.

\subsection{Comparation In-Depth for IVN-IDS}
To further present our advantages, the main characteristics and functionalities of the MKF-ADS are listed in Table ~\ref{tab1}. 
According to the modeling strategies and objectives, these works are divided into Statistical Learning (SL), Machine Learning (ML), and Deep Learning (DL). The majority of studies focus on DL-based and build different objectives, including supervised learning to identify known attacks, and anomaly-based to classify or predict unknown attacks. Intuitively, we find that early research based on SL and ML have similar characteristics that fixed feature extraction patterns ~\cite{muter2011entropy,studnia2018language, olufowobi2019saiducant} and worrisome performance~\cite{tanksale2019intrusion}. In contrast, supervised-based DL is substantially improved by rich feature extraction. For example, CNN extracts spatial features~\cite{song2020vehicle}, LSTM extracts temporal features~\cite{lo2022hybrid}, and the attention mechanism models valuable features~\cite{cheng2022stc}. Despite this, the shortcomings are evident, e.g., higher computation complexity and limited detection range. When the attention is transferred to anomaly classification, reconstruction modeling aims to attain the deviation score of normal messages~\cite{agrawal2022novelads, ashraf2020novel}. In general, this technique suffers from significant collection delays. Also, the one-class classifier is a bare-bones migration that always sacrifices the detection ability of known attacks~\cite{tariq2020cantransfer, zhang2023federated}.

In contrast, we tend to regard anomaly detection of IVN as a time-series prediction problem, due to lower collection delay and resisting more attacks. Traditional LSTM-based methods initially prove these advantages~\cite{taylor2016anomaly,qin2021application}. Subsequently, valuable feature extraction was considered to be equally important, but with the same complexity as the previous branch, i.e. the modeling of spatial-temporal features required more encoded components~\cite{zhang2019deep, sun2021anomaly}. From the perspective of knowledge modeling, it is a fact that SL and ML models only have simple and shallow criteria for detection. The existing DL-based works prove that spatial-temporal features are useful to IDS. Moreover, the contextual features extracted by the transformer-based model have comparable and even superior performance compared to the spatial-temporal ones~\cite{nam2021intrusion,alkhatib2022can}. However, we reject this way because of the exponential increase in complexity. 

Taking into account the complementarity of spatial-temporal and contextual features, the multi-knowledge fused is practical. Most importantly, for deploying IDS in real vehicle systems, lower computation complexity is required to satisfy real-time detection. However, only a limited amount of work has been performed on workstations for simulation testing~\cite{studnia2018language, song2020vehicle}. Some works~\cite{yang2021mth, olufowobi2019saiducant, cheng2022stc} have performed vehicle-level testing or real-time analysis. Hence, our proposed MKF-ADS constructs a cross-knowledge distillation framework that regards contextual knowledge as a teacher and spatial-temporal knowledge as a student. We hope that the proposed model can capture subtle deviations through enriching knowledge and then identify potential threats. Notably, we introduce fine-pruning to modeling spatial-temporal knowledge, and the reduced capabilities are instructed by contextual knowledge, which is a guarantee for real-time detection. Also, robustness validation is presented by fine-tuning the proposed MKF-ADS.

\section{Preliminaries}\label{sec3}
In this section, we provide a brief overview of vehicular communication.  Thereafter, we analyze in-vehicle vulnerability in-depth and present some attack scenarios. Also, the design intuition is discussed based on the observation of the collected CAN logs.
\subsection{Control Area Network Bus}

\begin{figure}
    \centering
    \includegraphics[width=1\linewidth]{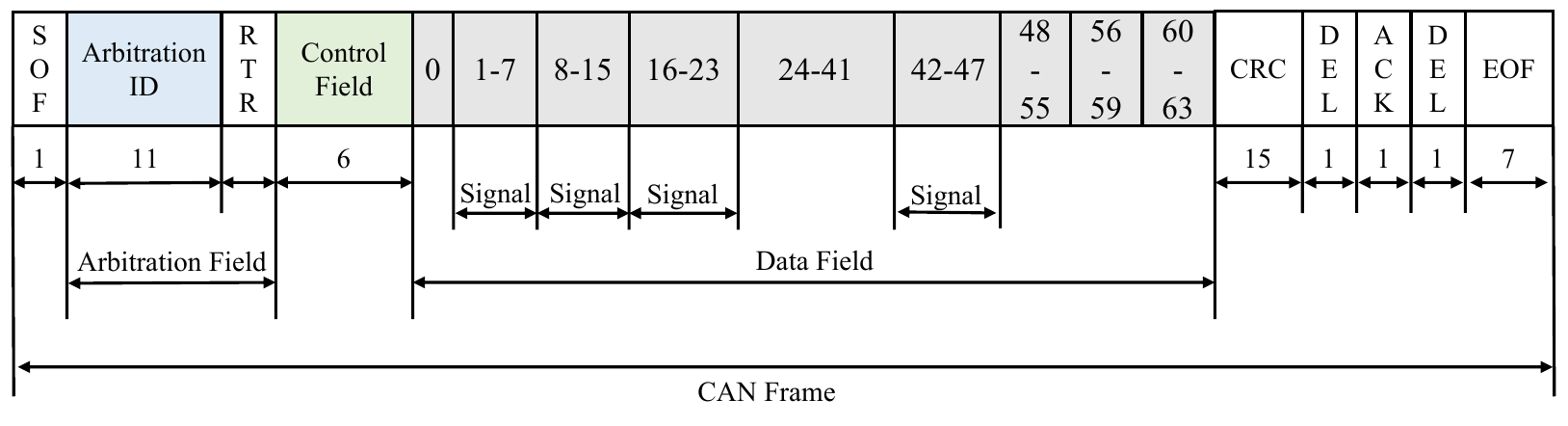}
    \caption{The standard structure of CAN frame. In the data field, each ID's signal is translated by the "READ" method. For instance, we extracted four effective physical signals on the '0x260' ID, including $1^{th}-7^{th}$, $8^{th}-15^{th}$, $16^{th}-23^{th}$, and $42^{th}-47^{th}$.}
    \label{fig:1}
\end{figure}
Control Area Network (CAN) bus is a de facto communication standard for vehicular networks, which divides ECUs into different domains, e.g., safety domain, power train domain, and comfort domain. Each of these communication sub-networks is responsible for different key tasks such as automatic crash notification, vehicle diagnostics, and infotainment services ~\cite{sodhro2020ai}. To achieve intelligent services, more external interfaces (e.g., communication via WIFI, Bluetooth, and 5G) are opened up, which also increases the potential vulnerability exposure. 

Notably, ECUs of each domain are interconnected with a central gateway and then send specific messages with a fixed periodic on the CAN bus. Fig.~\ref{fig:1} presents the standard structure of the CAN data frame. The arbitration mechanism (11 bits for CAN 2.0 A and 29 bits for CAN 2.0 B) decides what message to send based on priority. The control and data fields represent data transmission length and content. In this paper, we focus on the CAN ID and data fields because they always suffer from various cyber-attacks. 

\subsection{Attack Model}
In this study, we consider that an attacker can remotely compromise one or more ECUs via a wireless interface (e.g., a telemetries port ~\cite{cheng2022stc}) or physically (e.g., via OBD-II ~\cite{sun2023distillate}). Therefore, every ECU may become the potential entrance for the attacker to access the CAN bus and generate various attacks. The details of these attack models are summarized as follows:
\begin{itemize}[leftmargin=1em]
    \item \textit{\textbf{Denial of Service (DoS) Attack}}: The attack aims to refuse or interfere with the intended function of the CAN bus. The attacker usually sends a large number of redundant messages with higher priority to exhaust system resources and paralyze the functions. 
    \item \textit{\textbf{Fuzzy Attack}}: This attack will generate and send random CAN IDs and data frames. The randomly generated CAN ID ranges from 0x000 to 0x7ff, including the ones initially extracted from the vehicle.
    \item \textit{\textbf{Suspension Attack}}: This attack will try to compromise some ECUs in the CAN bus system and stop them from sending any CAN messages.
    \item \textit{\textbf{Replay Attack}}: This attack hijacks a valid message at a specific time, which will be transmitted later. The attacker usually uses it to obtain illegitimate authority.
    \item \textit{\textbf{Spoofing Attack}}: This attack first sniffs valid messages at a specific ECU and then impersonates compromised ECU by simulating their message transmission frequencies, while the related data contents are usually forged.
    \item \textit{\textbf{Masquerade Attack}}: This attack is more insidious as the legitimate messages will be deleted before injecting a fabricated message. 
\end{itemize}

These attack mechanisms are consistent with the work~\cite{sun2021anomaly}, categorized by message injection and falsification. On the one hand, message injection (e.g., DoS, Suspension, and Replay) is usually related to message frequency or statistical message sequence regarding CAN ID, thus proposed IDS should be sensitive for temporal knowledge. On the other hand, message falsification (Fuzzy, Spoof, and Masquerade) is generally correlated to content, thus proposed IDS should be effective for spatial knowledge. Moreover, context knowledge, i.e., semantic features, are more robust, which can make up for the knowledge modeling. Therefore, MKF-ADS caters to these requirements, so improving effectiveness and efficiency for detecting malicious messages.

\subsection{Design Intuition}
\begin{figure}
    \centering
    \includegraphics[width=1\linewidth]{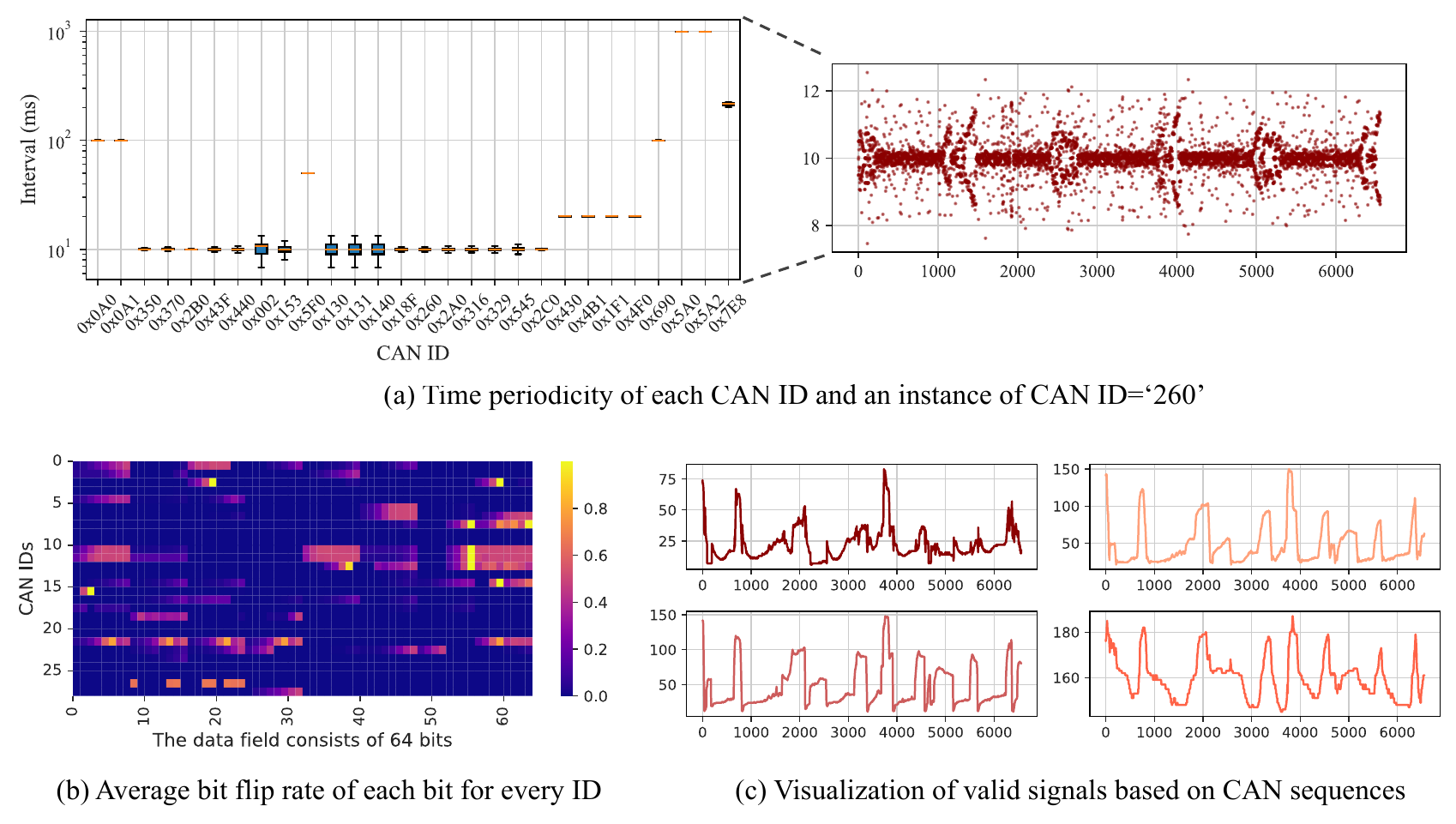}
    \caption{Illustration of the design intuition for knowledge modeling, of which (a) presents the frequency of CAN ID and an instance; (b) is the bit flip rate of each CAN ID; and (c) visualization of valid signals based on CAN sequences.}
    \label{fig:2}
\end{figure}

Indeed, feature modeling based on valuable knowledge is the cornerstone of ADS performance improvement. In this paper, we argue that temporal, spatial, and contextual knowledge are key to designing feature extractors. Existing works have demonstrated the importance of temporal~\cite{taylor2016anomaly, qin2021application} and spatial features~\cite{song2020vehicle, xun2021deep, lo2022hybrid} as well as spatial-temporal features~\cite{cheng2022stc}. We thus extend the pipeline on prediction-based work to further improve detection capability as well as guarantee efficiency.

In Fig.~\ref{fig:2}(a), CAN sequences send messages strictly following predefined frequencies (e.g., CAN ID=`260' has a fixed frequency of 10 ms), which requires our ADS to capture the valid temporal feature both on long-time and short-distance. Empirically, feature components are generally composed of recurrent neural networks (RNNs) with temporal capture capabilities. Besides, we observe dramatic changes of bit flip in local regions from Fig.\ref{fig:2}(b), which require our ADS to capture significant spatial features on all extracted signals. Generally, convolutional neural networks (CNNs) are fused on the pipeline. Last but not least, semantic features extracted by transformer-based components are currently the best knowledge for ADS. The bit flip rate in Fig.\ref{fig:2}(b) proves that the contextual relationships of signals vertically and different signal trends in Fig.\ref{fig:2}(c) demonstrate the interactions of valid signals horizontally. Although it is already well ahead in terms of performance, the complexity is higher~\cite{alkhatib2022can, nam2021intrusion}.

To this end, we proposed MKF-ADS which also utilizes a CNN-LSTM paradigm to model spatial-temporal knowledge.  Differently, parameters are significantly reduced to ensure efficiency. Meanwhile, the feature dimensional is dropped by the ``READ" algorithm. For instance, the CAN ID=`0x260' has four valid signals, as shown in Fig.~\ref{fig:1}. Moreover, we regard semantics as high-level knowledge with the ability to guide in order to empower spatial-temporal knowledge through knowledge distillation. MKF-ADS not only ensures the richness of knowledge but also keeps the computational complexity low.

\section{METHODOLOGY}\label{sec4}
\begin{figure*}[t]
    \centering
    \includegraphics[width=1\linewidth]{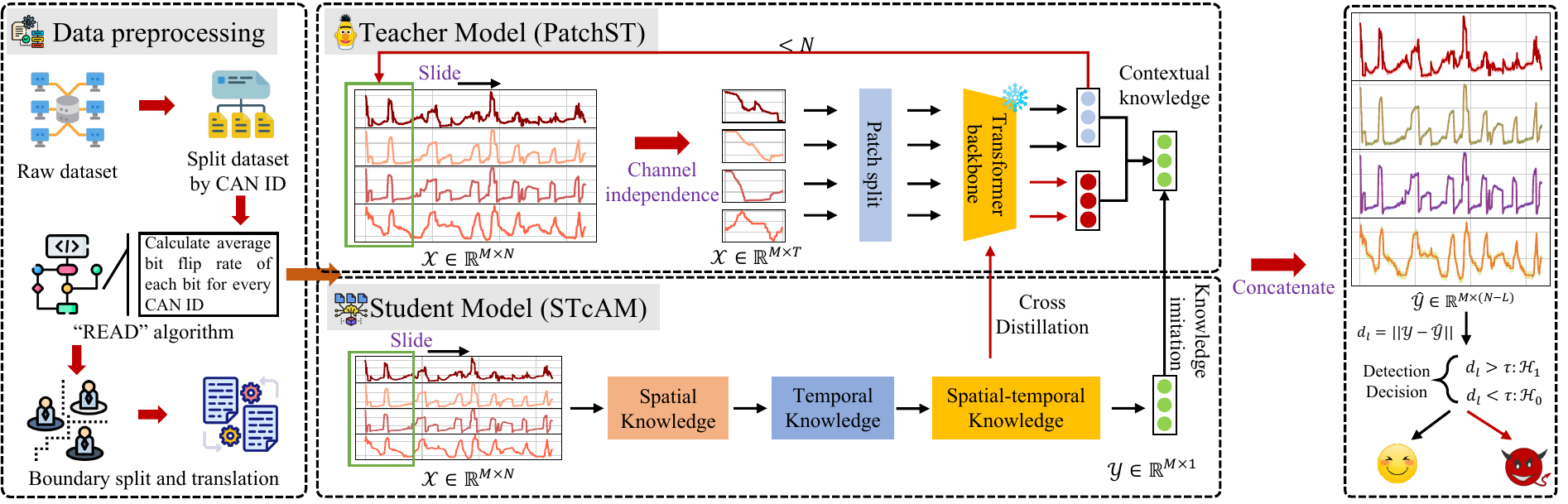}
    \caption{Overview of the proposed MKF-IDS.}
    \label{fig:3}
\end{figure*}

In this section, we explain how the MKF-ADS works, consisting of the spatial-temporal knowledge module (STcAM), contextual knowledge module (PathcST), and multi-knowledge fused module. The model design is shown in Fig. ~\ref{fig:3}.

\subsection{Overview}
The proposed MKF-ADS is devoted to advancing the comprehensive performance of in-vehicle ADS for the predictive paradigm. The improvements are distributed all over the pipeline, consisting of four parts as follows.
\begin{itemize}[leftmargin=1em]
    \item \textbf{Data preprocessing}. The physical payload is analyzed according to the bit flip rate and boundary decision algorithm, aiming to reduce the dimensionality of the data and improve the training efficiency of the model. 
    \item \textbf{STcAM}. The preprocessed data are fed into the STcAM to realize the spatial and temporal knowledge extraction. We also make it express significant timestamps using the soft attention mechanism.
    \item \textbf{PatchST}. To complement the body of knowledge, we extract fine-grained contextual knowledge of independent signals using the patch-based transformer model.
    \item \textbf{Cross Knowledge Distillation}. Aiming to build fused knowledge without increasing complexity, we introduce a cross-knowledge distillation framework to enhance the performance of the model. Therein, STcAM learns spatio-temporal knowledge maximally and continuously enters inside PatchST to mimic its contextual knowledge.
\end{itemize}
Overall, we can obtain an optimal threshold when MKF-ADS is trained until convergence. If the deviation between the predicted value and the received value exceeds the threshold, the CAN bus is considered to be compromised.

\subsection{Data preprocessing}\label{4.2}
The proposed model was first evaluated on the car hacking for intrusion detection dataset \footnote{https://ocslab.hksecurity.net/Datasets/CAN-intrusion-dataset} from the hacking and countermeasure research lab (HCRL). The dataset was collected via the OBD-II port connected by real-world vehicles. In the dataset, the crucial information for CAN messages (e.g., Timestamp, CAN ID, DLC, and DATA[0]-DATA[7]) is saved. The data pre-processing is an essential factor for DL-IDS because only if a model focuses on the significant areas for messages could the model have excellent performance. In this paper, we follow the prediction-based paradigm to execute data preprocessing, as shown in Fig.~\ref{fig:3}. Specifically, we split the payload sequence of CAN messages with the same ID from the raw dataset. Thereafter, we extract significant physical signals based on the ``READ'' algorithm and boundary split algorithm~\cite{marchetti2018read} without knowledge of the CAN communication protocol. For instance, the valid signal for CAN ID='260' is reduced from the original 64 bits to 41 bits. Next, we introduce a feature generator, which regards continuous raw CAN messages as multivariate time series and generates a discrete sub-sequence to be used as an input of MKF-ADS. The feature generator has two contributions: i) using a sliding window $T (T \in N | T > 10)$ to aggregate the recently arrived traffic and then produce a two-dimensional (2-D) vector from it; ii) translating the 2-D vector as a decimal format. In order to obtain fast convergence, all features were linearly normalized and calculated as follows: 
\begin{equation}
    \label{eq1}
    X = \frac{X - X_{min}}{X_{max}-X_{min}},
\end{equation}
Note that we only considered the CAN ID of crucial functional areas as they have fixed communication frequency (e.g., 10 ms). Hence, 17 IDs are filtered out of the raw dataset containing 28 IDs. After preprocessing, we parameterize MKF-ADS as $\mathcal{M}_\theta = \mathcal{M}_\theta(X, Y, ID)$, where $X(f_{1,\cdots, M})=\{X(f_1), X(f_2), \cdots, X(f_M)\}\in \mathbb{R}^{M\times T}$ is a training instance, $X(f_i)$ is a i-th physical signal, containing the $t$ observation set by the sliding window. $Y = \{y_1, y_2, \cdots, y_s\} \in \mathbb{R}^{M\times 1} $ is the corresponding label, $\mathcal{M}$ is a nonlinear mapping function. 

To evaluate the robustness of the proposed model, the Real ORNL Automotive Dynamometer (ROAD) CAN intrusion datasets \footnote{https://0xsam.com/road/} were used, which contain more comprehensive data on attacks such as fuzzing attacks, targeted ID fabrication, and masquerade attacks. In contrast to HCRL, this dataset is performed on a variety of scenarios and target IDs, as well as expanding on the attack types. Data pre-processing of it also implemented with the same standard via the feature generator. 

\subsection{Proposed MKF-ADS}
MKF-ADS is an anomaly detection model, consisting of STcAM, PatchST, knowledge fused, and an anomaly detection module, which analyses the deviation between the predicted sequence and the input sequence to determine a reasonable threshold. The underlying hypothesis is that normal statistical feature behavior does not change abruptly and conversely, abnormal behavior can lead to dramatic changes in statistical features and deviations from threshold criteria. The workflow in this paper consists of a training phase and a testing phase, running on a workstation and in a real in-vehicle environment, respectively.
\begin{figure}[t]
    \centering
    \includegraphics[width=1\linewidth]{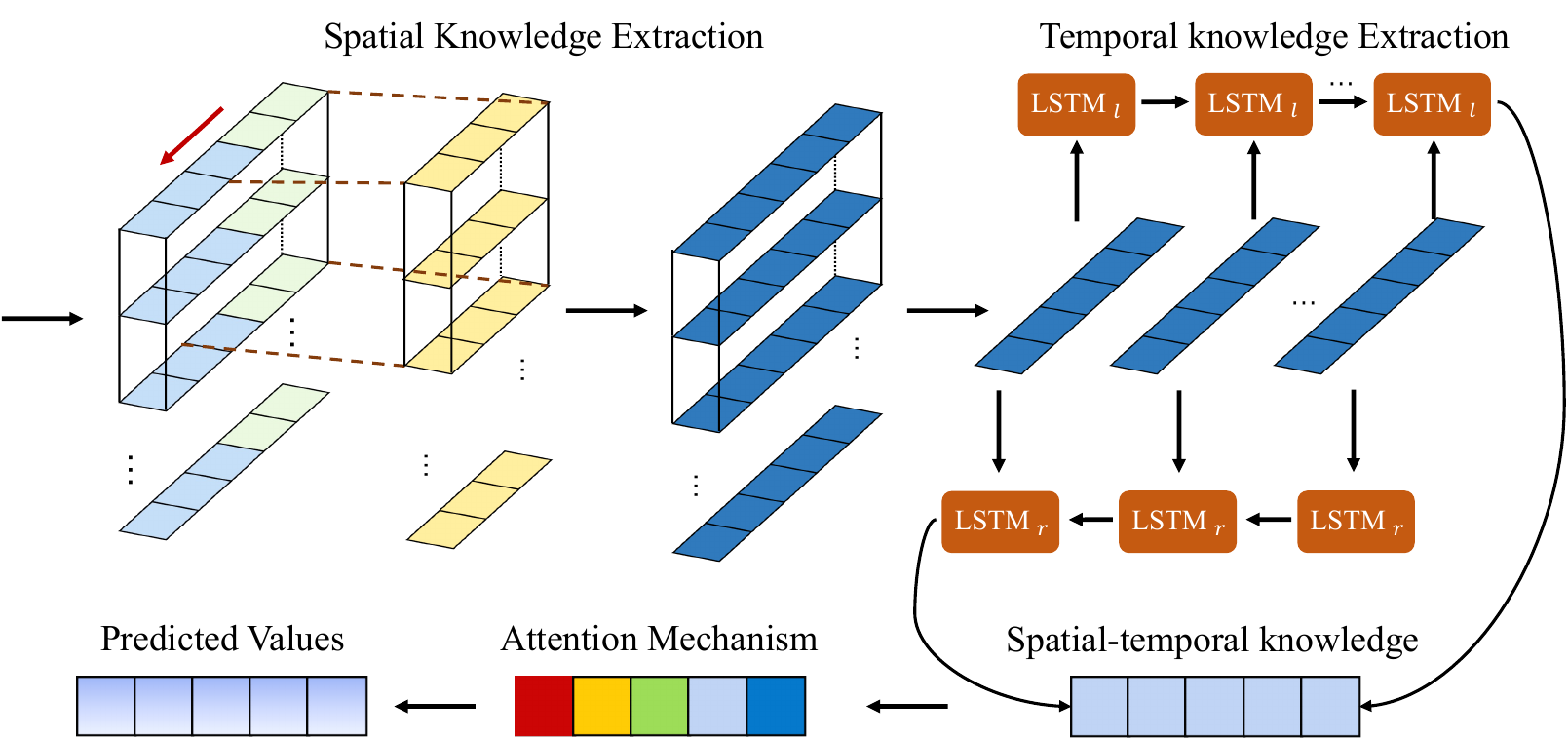}
    \caption{Structure of STcAM component.}
    \label{fig:4}
\end{figure}

\subsubsection{\textbf{STcAM Module}}
The module is an important implementation that was originally designed to be sufficiently lightweight compared to previous works under a guarantee for detection performance. Given the multivariate time series of the i-th time window, we use a one-dimensional CNN layer with $w$ filters as the first component of the module to extract spatial knowledge, calculated as follows:
\begin{equation}
    s_{t}^j = h_t^j(f)*x_t^j(f) = \sum_{m=0}^{N} h_i^j(f-m)x_t^j(m),
\end{equation}
where $s_{t}^j$ denotes the $t$-th spatial feature for convolution calculation.
$h_t^j(f)$ and $x_t^j(f)$ denote the convolution kernel and input, respectively, $f$ is the kernel size with $ 3 \times 1$, and $t$ denotes the index of the data in the sequence. To nonlinear the features, the rectified linear unit (ReLU) function is utilized to implement the retention of negative features and the discarding of negative features, calculated as follows:
\begin{equation}
    z(s) = s^+ = max(0, s).  
\end{equation}

Subsequently, it will continue to capture temporal knowledge based on BiLSTM. The purpose is to maintain the timing dependence in the window size $w$ and to guarantee the timing corruption due to window sliding, i.e., to learn the timing characteristics in both directions. The core of LSTM implementation for each direction in the module is to maintain the information via cell state. Hence, the latent representation $h$ with spatial-temporal relationship is calculated as follows:
\begin{equation}
    h_t = \mathrm{o}_t * \sigma_1(c_t), 
\end{equation}
where $o_t$ and $c_t $ represent the output gate and cell state activation value. The cell gate depends on the retention of information from the previous moment $c_{t-1}$ by the forget gate $\mathrm{f_t}$ and the update of the current state information $\tilde{c}_t$ by the input gate $\mathrm{i_t}$, calculated as follows:
\begin{equation}
    c_t = (f_t * c_{t-1}) + (i_{t}*\tilde{c}_t).
\end{equation}

Moreover, information transmission with three gates: forget gate $\mathrm{f}$, input gate $\mathrm{i}$, and output gate $\mathrm{o}$ on current moment $t$, is calculated as follows: 
\begin{equation}
    \begin{aligned}
        f_t & = \sigma_2 (W_f[h_{t-1}, x_t], b_f),\\
        i_t & = \sigma_2 (W_i[h_{t-1}, x_t], b_i),\\
        o_t & = \sigma_2(W_o[h_{t-1}, x_t], b_o), \\
        c_t & = \sigma_1 (W_c[h_{t-1}, x_t], b_c),
\end{aligned}
\end{equation}
where $W$ and $b$ are weights of the input and bias parameters required to be learned during the training of the MKF-ADS. Also, the activation functions are calculated as follows:
\begin{equation}
    \begin{aligned}
    \sigma(\cdot)_1 & = tanh(x) = \frac{e^{2x}-1}{e^{2x}+1}, \\
    \sigma(\cdot)_2 &= sigmoid(x) = \frac{e^x}{e^x+1}.
    \end{aligned}    
\end{equation}

After obtaining the forward spatial-temporal knowledge $(\overrightarrow{h_1}, \cdots, \overrightarrow{h_t})$ and backward spatial-temporal knowledge $(\overleftarrow{h_1}, \cdots, \overleftarrow{h_t})$, we concatenate and obtain the final output as $h_i=\left[\overrightarrow{h_i} ; \overleftarrow{h_i}\right]$. To lightweight, all components of spatial-temporal are parameters-shared. Moreover, there is an indispensable aim that focuses on significant time steps. Hence, we introduce the soft attention mechanism (AM) to add the adaptive weight for spatial-temporal representation, calculated as follows:
\begin{equation}
    \alpha_i = \frac{exp(h_i)}{\sum^{t-1}_{i=1}exp(h_i)},
\end{equation}
where $\alpha_i$ is the conditional probability distribution, calculated by the softmax function. Finally, the attention vector of spatial-temporal knowledge is presented as follows:
\begin{equation}
    h^* = \sum^{t-1}_{i=1}\alpha_i h_i.
\end{equation}
To prevent overfitting, the dropout layer is introduced. Then, the output vector is fed into a dense layer to predict the future time step. The STcAM component is depicted in Fig. ~\ref{fig:4}.


\subsubsection{\textbf{PatchST Module}}
Transform-based high-level context knowledge is a state-of-the-art approach for anomaly detection in multivariate temporal sequences, due to its long-range modeling capability. Inspired by~\cite{nie2022time}, we introduce PatchST and regard it as complementary to spatial-temporal knowledge, relaxing the restriction that makes it difficult to use complex models for resource-constrained IVNs~\cite{alkhatib2022can, guo2021logbert}.
\begin{figure}[t]
    \centering
    \includegraphics[width=1\linewidth]{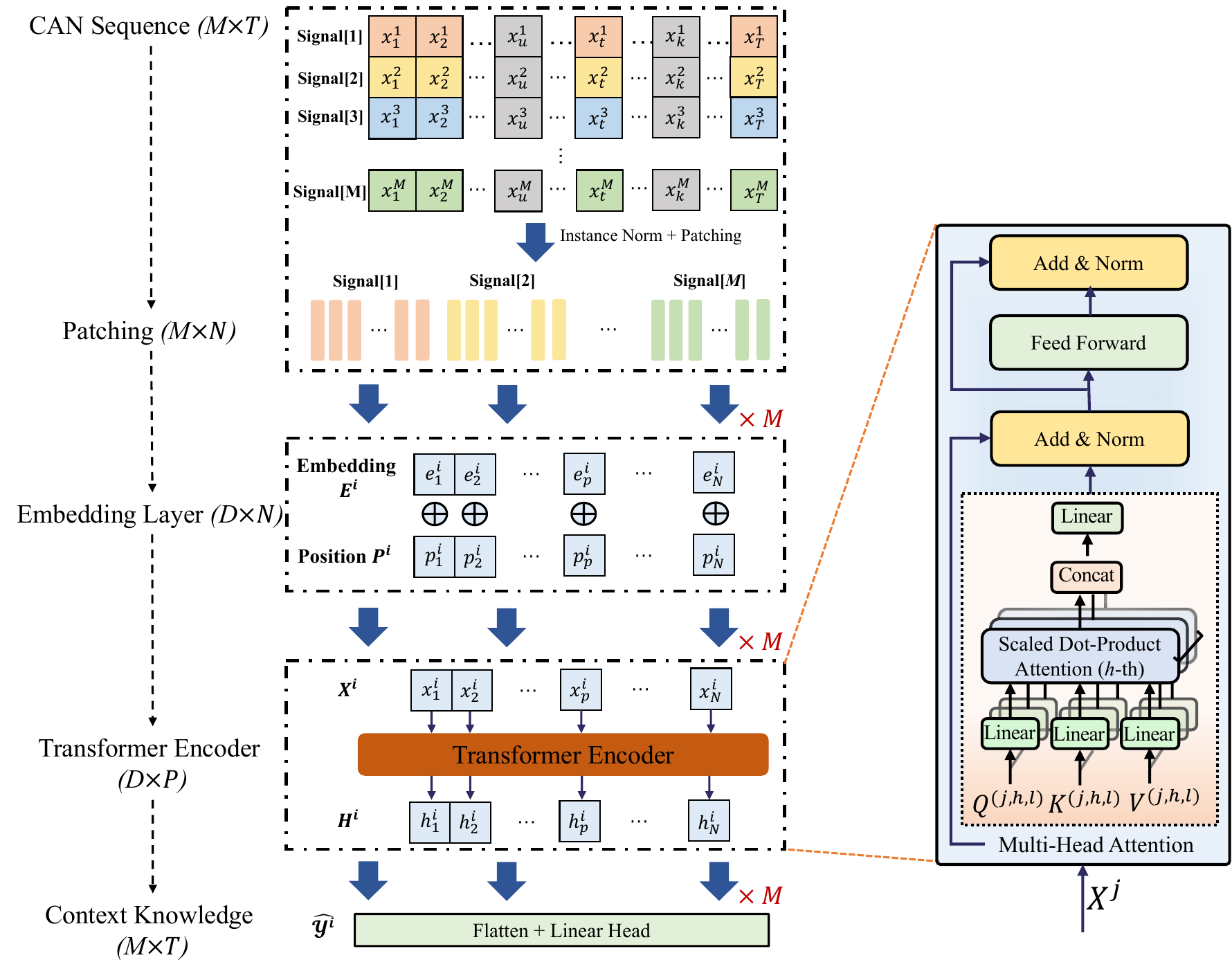}
    \caption{Structure of PatchST component.}
    \label{fig:5}
\end{figure}

Given a collection of multivariate time series instances with lookback window $T$: $(x_1,\dots, x_T) $ where each $x_t$ at time step $t$ is a vector of dimension $M$, we extract the context knowledge to forecast the future physical signal. The PatchST is illustrated in Fig.~\ref{fig:5}. We first split CAN sequence $X\in\mathbb{R}^{M\times T}$ into $M$ univariate series $x^{(i)}\in \mathbb{R}^{1\times T}$, whose length $T$ starting at time index 1 to slide window size $T$. Each univariate series is fed independently into the transformer backbone according to the channel-independence setting. Then the Transformer-based model will forward prediction results $\hat{y}^{(i)}$.

\textbf{Patching.} Given a univariate time series $x^{(i)}$, we first divide it into patches $x^(i)_{p} \in \mathbb{R}^{P\times N}$, where  $P$ is the patch length, $N$ is the number of patches, $N=\left\lfloor\frac{(L-P)}{S}\right\rfloor+2$ and $S$ is the stride of the nonoverlapping region between two consecutive patches. Here, the number of input tokens can reduce from $T$ to approximately $T/S$, thus the attention map is quadratically decreased by a factor of S. In other words, the proposed PatchST is sparse and focuses on the local and longer historical sequence simultaneously, which can significantly improve the forecasting performance.

\textbf{Embedding.} The patches are mapped to the Transformer latent space of dimension $D$ via a trainable linear projection $W_p \in \mathbb{R}^{D\times P}$, and a learnable additive position encoding $W_{pos} \in \mathbb{R}^{D\times N}$, calculated as follows:
\begin{equation}
    \begin{aligned}
        e_{t}^i &= W_ex_{p}^i + b_e,\\
    p_{t}^i &= W_{pos}x_{p}^i+b_{pos},
    \end{aligned}
\end{equation}
where $e_{t}^{i}$ and $p_{t}^i$ represents the embedding feature and positional embedding. Thus, the model input could present $x_d^i = e_t^i + p_t^i$, where $x_d^i \in \mathbb{R}^{D\times N}$ will be fed into Transformer encoder.

\textbf{Transformer encoder.} We use a vanilla Transformer encoder to extract context semantic knowledge, consisting of two sub-layers: a multi-head self-attention mechanism and a position-wise feed-forward network. Besides, a residual connection is utilized between two sub-layers, followed by layer normalization. For the $l$-th transformer encoder, each head $h=1,\cdots, H$ in multi-head attention transform $x_d^{(i)}$ into query matrices $Q_h^{(i)} = (x_d^{(i)})^TW_h^Q$, key matrices $K_h^{(i)} = (x_d^{(i)})^TW_h^K$, and value matrices $V_h^{(i)} = (x_d^{(i)})^TW_h^V$, where $W_h^Q, W_h^K \in \mathbb{R}^{D\times d_k}$ and $W_h^V \in \mathbb{R}^{D\times D}$. After that, attention output $O_h^{(i)} \in \mathbb{R}^{D\times N}$ is calculated as follows:
\begin{equation}
O_i^{(i)}=\sigma\left(\frac{Q_h^{(i)} K_h^{(i) T}}{\sqrt{d_k}}\right) V_h^{(i)},
\end{equation}
where $h$ represents the $h$-th self-attention, and $\sigma$ is the softmax function.

Further, all $O_h^{(i)}$ are concatenated by column and projected into representation space via a linear layer, which is the final multi-head attention result in $l$-th transformer encoder, calculated as:
\begin{equation}
    X_A^{(i, l)} = W^A[O^{(i, l)_1}, \dots, O^{(i, l)_h}, \dots, O^{(i, l)_H}],
\end{equation}
where $W^A$ is the attention output weight, and $H$ represents the total amount of the self-attention module. 

Next, the residual connection and layer normalization module are connected to each sub-layer of the transformer encoder. Residual connection is utilized to allow multi-transformer layers to be superimposed without degradation. Moreover, layer normalization has the competence to make the input of each neural under the same value of average and variance, to accelerate the network convergence. Hence, output representations are calculated as follows:
\begin{equation}
    X_{N}^{(i, l)} = L(X^{(i, l)} + X_A^{(i,l)}) = L(X^{(i, l)} + A(X^{(i,l)})),
\end{equation}
where $A(\cdot)$ is the output function of the multi-head attention module. $L(\cdot)$ is the layer normalization function, calculated as:
\begin{equation}
    L(X^{(i,l)}) = ReLU(\frac{g^m}{\sigma^m}(X^{(i,l)}-\mu^l)+b), 
\end{equation}
where $g$ and $b$ represent the gain and bias training parameters. $\mu^m = \frac{1}{N}\sum_{i=0}^N X^{(i,l)}$ and $\sigma^m = \sqrt{\frac{1}{N}\sum_{i=0}^N(X^{(i,l)}-\mu^l})^2$ are the average and variance of $m$-th layer. 

Although $X_A^{(i, l)}$ obtains the contextual relationships of the messages at each patch, it nevertheless performs a matrix multiplication, i.e. a linear transformation, and thus requires reinforcing the feature representation with a non-linear transformation. Hence, the position-wise feed-forward with two liner layers is applied to compute nonlinear attention features:
\begin{equation}
    X_F^{(i, l)} =W_{f2}\sigma( W_{f1}X_{N_1}^{(i, l)} + b_{f1})+b_{f2},
\end{equation}
where the $W_{f1}, W_{f2}$ and $b_{f1}, b_{f2}$ are the weight and bias of the module. $\sigma$ is the first layer activation function ReLU. Immediately, $X_F^{(i, l)}$ is fed into the second layer normalization module, thereby obtaining the final valuable context features $H^{(i, l)} = \{h_1^{(i, l)},\dots, h_p^{(i, l)}, \dots, h_N^{(i, l)}\}$ of $l$-th encoder. After forwarding through all transformer layers, a flatten layer with linear head is employed to obtain the prediction physical signal $\hat{y}^{(i)}$.

\subsection{Cross Knowledge Distillation}
There is an unreality suppose that use the transformer-based model for IVN-ADS. To achieve this capability, we introduce knowledge distillation to make STcAM mimic the PatchST knowledge. However, we observe that directly mimicking the teacher confronts the convergence conflict problem. To this end, we introduce cross-knowledge distillation, aiming that the STcAM can obtain a positive contribution from the PatchST.

For a given teacher-student pair, the predictions of the teacher and student can be represented as $p^t$ and $p^s$, respectively. Thereafter, suppose the intermediate output of STcAM $X_a^{(i)}$ (e.g., the output of attention mechanism), we deliver it to the teacher component, resulting in the cross-head predictions $\hat{p}^s$. Hence, the mimic knowledge process is described as follows: 
\begin{equation}
\mathcal{L}_{C K D}=K L\left(\hat{p}^s \| p^t\right)=\sum_{i=0}^N \hat{p}_i^s \log \frac{\hat{p}_i^s}{p_i^t},
\end{equation}
where KL represents the Kullback–Leibler (KL) divergence. The final optimal object controlled by hyperparameter ($\alpha$ and $\beta$) is calculated as follows:
\begin{equation}
    \begin{aligned}
    \mathcal{L} &= \alpha \mathcal{L}_{preds} +\beta \mathcal{L}_{CKD} \\
    &= \frac{1}{N}\sum_{i=0}^N(y_i-\hat{y}_i)^2 + \beta \sum_{i=0}^N \hat{p}_i^s \log \frac{\hat{p}_i^s}{p_i^t},
    \end{aligned}
    \label{eq8}
\end{equation}
where the $\mathcal{L}_{preds}$ is MSE function for online training phase. In the offline testing phase, we use Mean Absolute Error (MAE): $\frac{1}{m}\sum_{i=1}^{m}|y_i-\hat{y}_i|$ to evaluate the prediction performance. Note that the proportion of prediction deviation based on ground truth is determined by Mean Absolute Percentage Error (MAPE): $\frac{1}{m}\sum_{i=1}^m\frac{y_i-\hat{y}_i}{y_i}$. The proposed model is described in Algorithm~\ref{alg1}.
\begin{algorithm}[t]
\DontPrintSemicolon
  \SetAlgoLined
  \KwIn {\text{Training-loader} $X_{D}$, \text{Testing-loader} $X_{test}$; STcAM, PatchST}
  \KwOut {Anomaly detection result $res = []$, MKF-ADS;}
  \#\#Training Phase:\;
  \While{PatchST is not Convergence}{
    \For{$X \in X_{D}$}{
     $X_p = $Patching$(X)$\;
     $X_d = $Embedding$(X_p)$\;
     $\hat{y}$ = PatchST$(X_{d})$\;
     $\mathcal{L} =$ MAE$(y_i, \hat{y}_i)$\; 
     $\mathcal{L}.backward()$\;
    }
  }
  \While{MKF-ADS is not Convergence}{
    Initialization: PatchST.eval()\;
    \For{$X \in X_{D}$}{
     $p^t = $PatchST$(X)$\;
     $\hat{y}$,$X_a$ = STcAM$(X)$\;
     ${\hat{p}}^s$ = PatchST$(X_{a})$\;      
     $\mathcal{L} = \mathcal{L}_{preds}(y, \hat{y}) + \mathcal{L}_{CKD}({\hat{p}}^s, p^t)$ \;
     $\mathcal{L}.backward()$\;
    }
   }
   Calculate Threshold: $\tau$\ by MAPE; \;
    \#\#Testing Phase:\;
   \For{$X \in X_{test}$}{
       $\hat{y} = $MKF-ADS$(X)$\;
       $\mathcal{L} = MAE(y, \hat{y})$\; 
       res.append(metric$(\mathcal{L}, \tau))$
   }
  \caption{MKF-ADS}
  \label{alg1}
\end{algorithm}

\section{Performance Evaluation}\label{sec5}
In this section, we conduct an extensive evaluation of MKF-ADS for effectiveness and efficiency in preventing malicious attacks on vehicular networks.
\subsection{Experiment Setup}
\subsubsection{Hardware \& Software} To develop the proposed MKF-ADS, we use the PyTorch framework in Python to train the model. In the training phase, the model was carried out on Intel(R) Core (TM) i7-9500U CPU@3.6GHZ, 64 GB of RAM, and GPU RTX 3090. In the testing phase, the model's performance and effectiveness were evaluated using real automobiles with NVIDIA Jetson AGX Xavier after malicious messages with the same specification were first injected through CAN Test software (16 GB). 
\subsubsection{Metrics} 
In the publicly available dataset, we considered each abnormal sequence as a positive case, while the true sequence was considered a negative case. Accordingly, we calculated four statistical metrics: true positive (TP), true negative (TN), false positive (FP), and false negative (FN), respectively. Thereafter, we evaluated the precision (P), recall (R), and F1-Score of the model. 
\begin{equation}
    P = \frac{TP}{TP+FP}
\end{equation}
\begin{equation}
    R = \frac{TP}{TP+FN}
\end{equation}
\begin{equation}
    F1 = \frac{2\times P\times R}{P+R}
\end{equation}
where high precision is the security basis of vehicle network intelligent services, which could identify malicious attacks as much as possible, and a stable recall rate can prove that the model will not miss negatives, otherwise it will cause serious security risks. Also, F1 represents a compromise between precision and recall, which is ideal for unbalanced CAN message data.

Further, both error rate (ER) and false alarm rate (FAR) are reported, which are significant measures of detection system performance, calculated as:
\begin{equation}
    ER = \frac{FP+FN}{TP+TN+FP+FN}
\end{equation}
\begin{equation}
    FAR =\frac{FP}{FP+TN}
\end{equation}
where ER donates the number of incorrect classifications as a percentage of the total sample. 
A lower FAR represents high-quality service from the IDS without constant false alarms.
\subsubsection{Implementation details} We leverage the grid-search to obtain the best hyperparameters (e.g., slide window $T \in \{8, 16, 32, 64\}$, filter size $f \in \{8, 16, 32, 64\}$, hidden units $h \in \{4, 8, 16, 32\}$, dropout is $0.2$, and batch size $b \in \{128, 256, 512, 1024\}$). For the teacher model (PatchST), all CAN IDs employ patch length $P=4$, stride $S=1$ on 8 embedding dimensional and 2 heads of the transformer-based model. Moreover, the optimizer is Adam, the learning rate is 1e-2, and epochs are 3000 with early stopping.
\subsubsection{Baseline Models}
We compared our MKF-ADS method with four baseline methods as follows: 
\begin{itemize}[leftmargin=1em]
    \item \textbf{LSTM-P:} This model uses the LSTM network to extract temporal knowledge and then predict 64 bits physical value for the next time step~\cite{taylor2016anomaly}.
    \item \textbf{LSTM-E:} Unlike LSTM-P, this model extracts temporal knowledge and then predicts decimal physical value with improved optimization method~\cite{qin2021application}. 
    \item \textbf{DeepConvGRU:} This model uses a deeply separable convolutional layer and two extended RNN layers to extract spatial-temporal knowledge~\cite{zhang2019deep}.
    \item \textbf{CLAM:} This model is a baseline for MKF-ADS, of which the prediction results only refer to spatial-temporal attention knowledge~\cite{sun2021anomaly}.
\end{itemize}
\begin{figure}[t]
    \centering
    \includegraphics[width=1\linewidth]{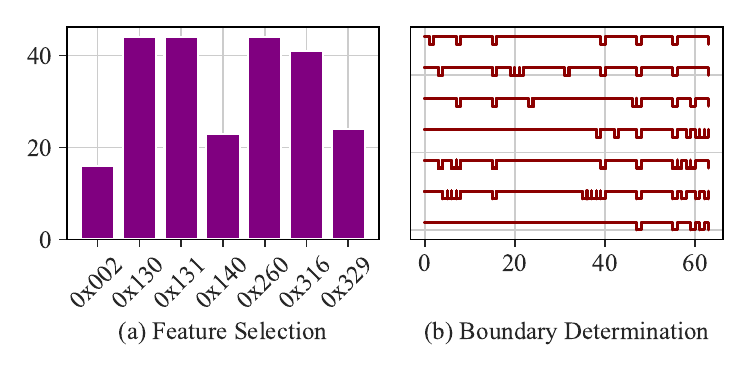}
    \caption{CAN IDs are evaluated on the  MKF-ADS, where (a) represents the feature selection based on the ``Read" algorithm, and (b) represents the boundary determination for candidates CAN IDs.}
    \label{fig:id}
\end{figure}
\begin{figure}[!t]
    \centering
    \includegraphics[width=1\linewidth]{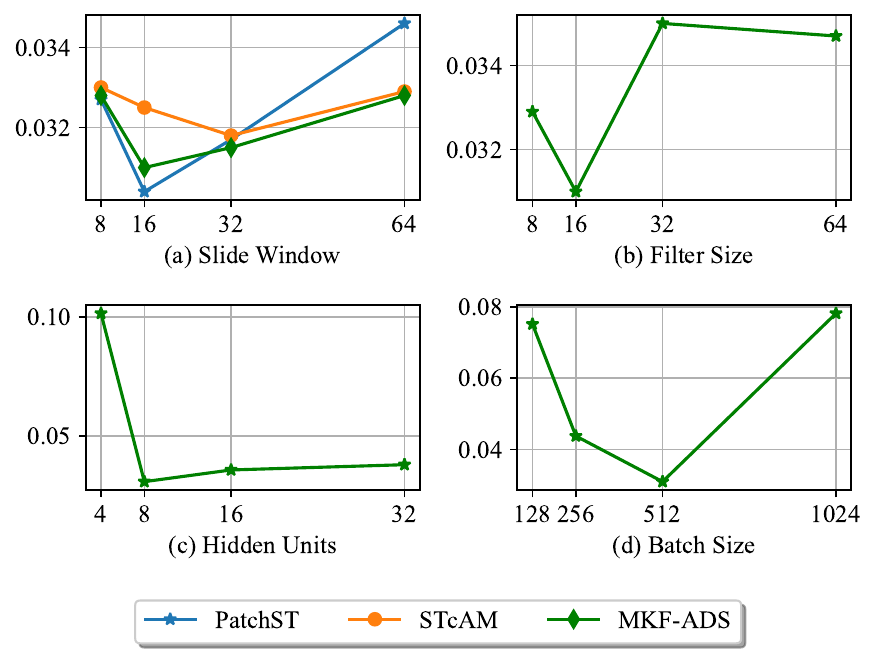}
    \caption{Determination of crucial hyperparameters based on grid search.}
    \label{fig:6}
\end{figure}
\subsection{Detection Performance}

To evaluate the performance of the MKF-ADS, we build a multivariate time series for the same CAN ID. As described in dataset preprocessing in Section~\ref{4.2}, we first choose crucial features based on the ``Read" algorithm. As shown in Fig.~\ref{fig:id} (a), we filtered 7 out of 28 CAN IDs for evaluating MKF-ADS based on the rule that the number of features is greater than 15 and the periodicity is around 10 ms. Fig.~\ref{fig:id} (b) represents different signal channels for candidate CAN IDs, calculated by the boundary determination algorithm. For the offline training, we use 80\% of the dataset for training and 20\% for testing. Note that if not specifically mentioned, the main evaluation is on CAN ID = `0x260'.

In order to optimize MKF-ADS, we determined the optimal hyperparameters through a grid search, as shown in Fig.~\ref{fig:6}. For prediction-based ADSs, the sliding window determines how long historical information acts on the predicted reference. We find that shorter retrospectives do not provide a better prediction of either contextual or spatial-temporal knowledge. Similarly, longer historical information contains noisy signals that make the prediction performance drop dramatically. To ensure the accuracy of the prediction, we finally set the size of the sliding window to 16. Moreover, the filter size and hidden cells of the proposed MKF-ADS are 16 and 8, respectively, presenting the lowest MAE while reducing a larger number of parameters. In contrast, performance will suffer if the parameters are continuously reduced. Furthermore, the continued addition of learnable parameters is not a positive contribution to prediction. Importantly, batch size affects the convergence of the model at the rated learning rate, as unreasonable settings will be detrimental to the cross-fertilization of knowledge.
\begin{figure*}[t]
    \centering
    \includegraphics[width=1\linewidth]{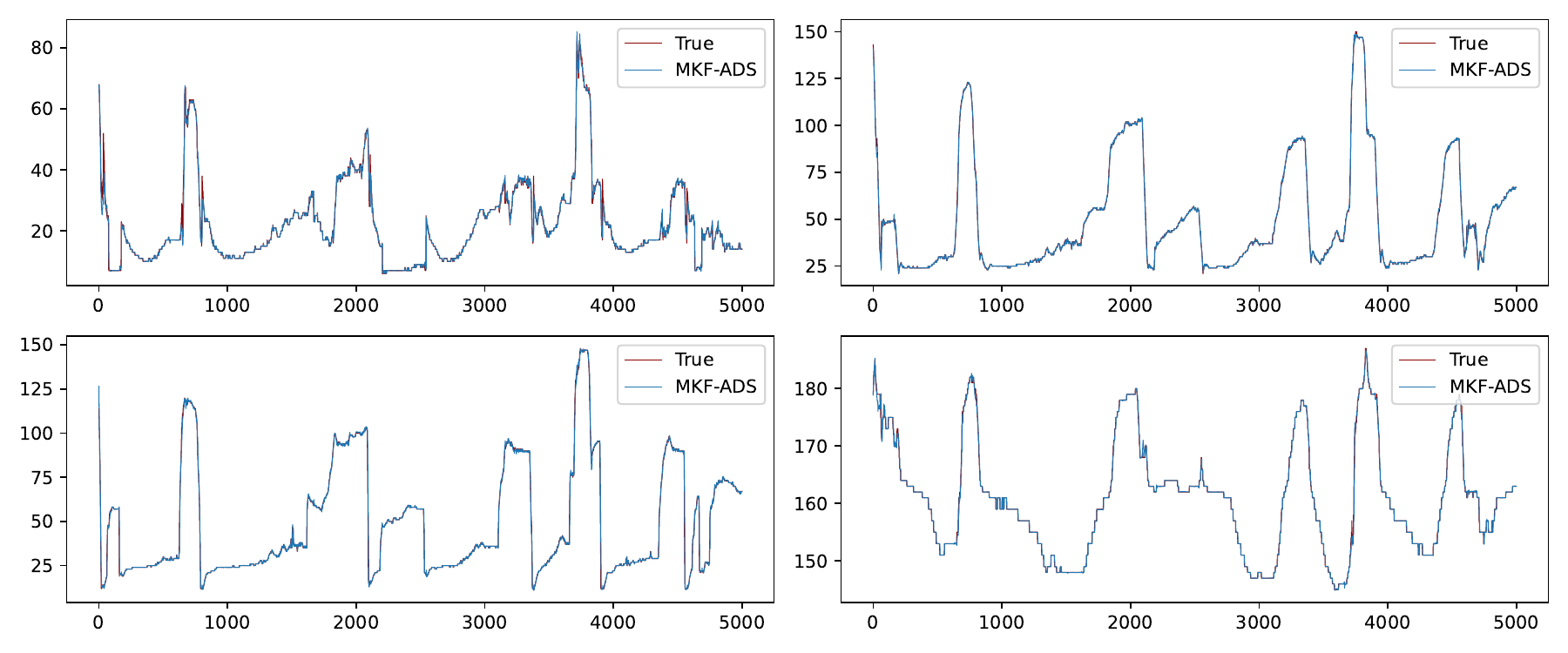}
    \caption{The prediction results of the MKF-ADS model. The x-axis represents the CAN sequence, and the y-axis represents the value of the signal. The red line is the ground truth and the blue line is the prediction result of the MKF-ADS.}
    \label{fig:7}
\end{figure*}
According to the chosen hyperparameters, MKF-ADS will converge quickly and obtain the global optimal solution. The prediction results of MKF-ADS are shown in Fig.~\ref{fig:7}, where the horizontal axis represents the sequences of CAN messages, and the vertical axis represents the signal value. One of the most intuitive results is that the predicted values closely match the actual values, which proves that MKF-ADS effectively captures the trend of the CAN sequence. Compared to the work~\cite{sun2021anomaly}, the proposed model maintains better prediction results in fine-grained variations of the signal. We believe that this is where knowledge fusion plays a positive role. 

\begin{table}[t]
    \centering
    \caption{prediction deviation of the model compared with baseline.}
    \label{tab2}
    \begin{tabular}{p{2cm}p{1.5cm}p{1.5cm}p{1.5cm}}
    \toprule
    Model & MAE & MAPE & RMSE \\ \midrule
    LSTM-P~\cite{taylor2016anomaly}&  0.3824 & - & 0.7350\\
    LSTM-E~\cite{qin2021application}& 0.0565 & 0.6939 & 0.1387 \\ 
    DeepConvGRU~\cite{zhang2019deep} & 0.0492 & 0.7355 & 0.1082\\
    CLAM~\cite{sun2021anomaly} & 0.0391 & 0.2547 & 0.0905\\  \midrule
    PatchST & 0.0314 & 0.1779 & 0.0858\\
    STcAM &  0.0325 & \textbf{0.1855} & 0.0955\\
    MKF-ADS & \textbf{0.0305} & 0.1992 & \textbf{0.0832}\\ \bottomrule
    \end{tabular}
\end{table}

Indeed, a well-trained anomaly detection model should maintain a low error with respect to normal messages. Table~\ref{tab2} presents the prediction performance compared to the baseline, where the knowledge is incremental. Evolving from temporal knowledge, spatial-temporal knowledge, and spatial-temporal attention knowledge to the proposed knowledge fusion framework, our model outperforms the other models. Interestingly, the LSTM-P model has the worst performance and the LSTM-E model follows because of the lack of knowledge. However, when the redundant features are removed, the MAE and RMSE of LSTM-E decrease to 0.0565 and 0.1387. Thereafter, the MAE and RMSE of DeepConvGRU decreased slightly when the spatial features were incorporated into the model, which are 0.0492 and 0.1082, respectively. CLAM is the current state-of-the-art model because the attention mechanism makes it easier to converge the model and jumps out of the local minima, resulting in MAE and RMSE of only 0.0391 and 0.0905. Moreover, the MAPE shows the same trend, decreasing from 0.6939 to 0.2547. In contrast, the proposed STcAM component continues to reduce the prediction error under a reduced model capacity, with only 0.0325 on the MAE, and 0.1855 on the MAPE. Also, the PatchST model implements the modeling of contextual knowledge with the lowest error. Thus, the proposed MKF-ADS achieves a fabulous prediction performance, with only 0.0305 on the MAE.

\begin{figure*}[t]
    \centering
    \includegraphics[width=1\linewidth]{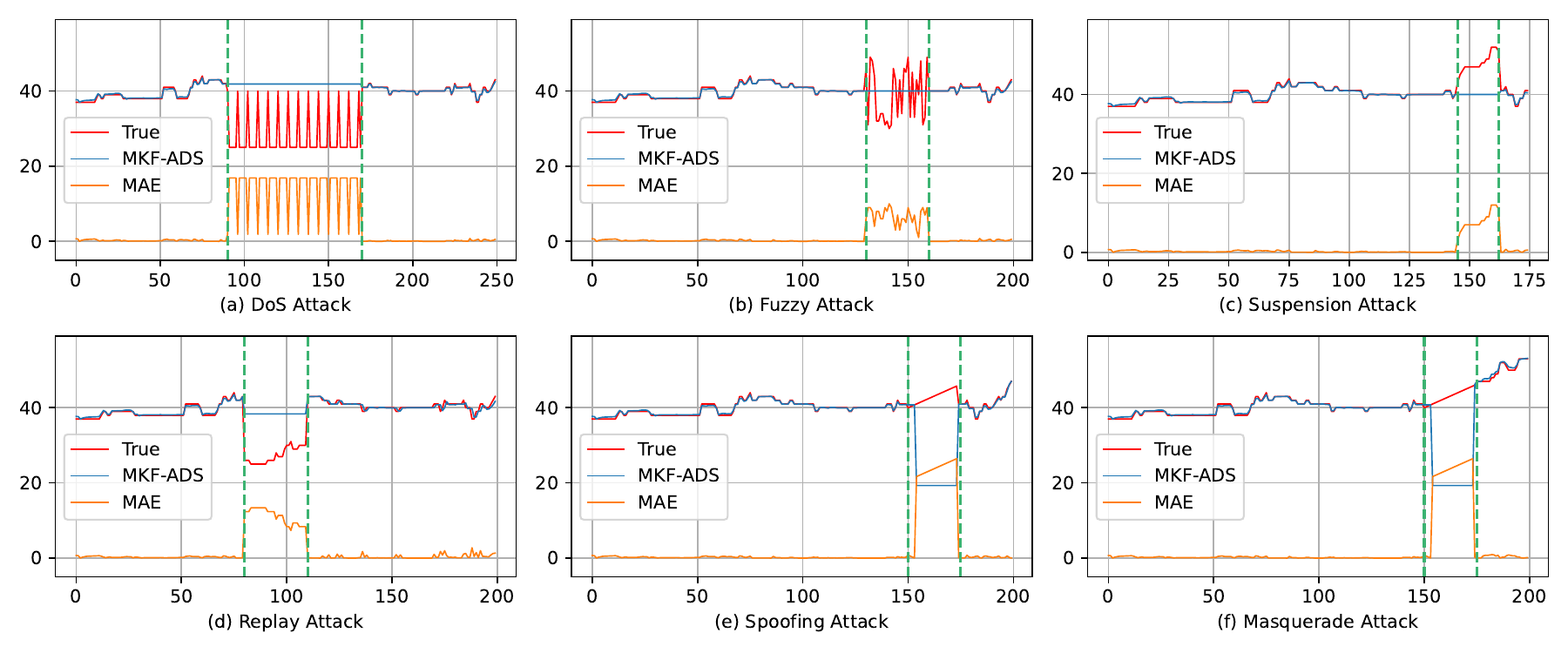}
    \caption{Visualization of the detection performance of coarse-grained attack, i.e., the attacker injects the abnormal deviates far away from the normal value. The red line is the ground truth of the CAN sequence on the CAN bus, the blue line is the predicted value, and the yellow line is the anomaly score, calculated by MAE.}
    \label{fig:8}
\end{figure*}

\begin{figure*}[t]
    \centering
    \includegraphics[width=1\linewidth]{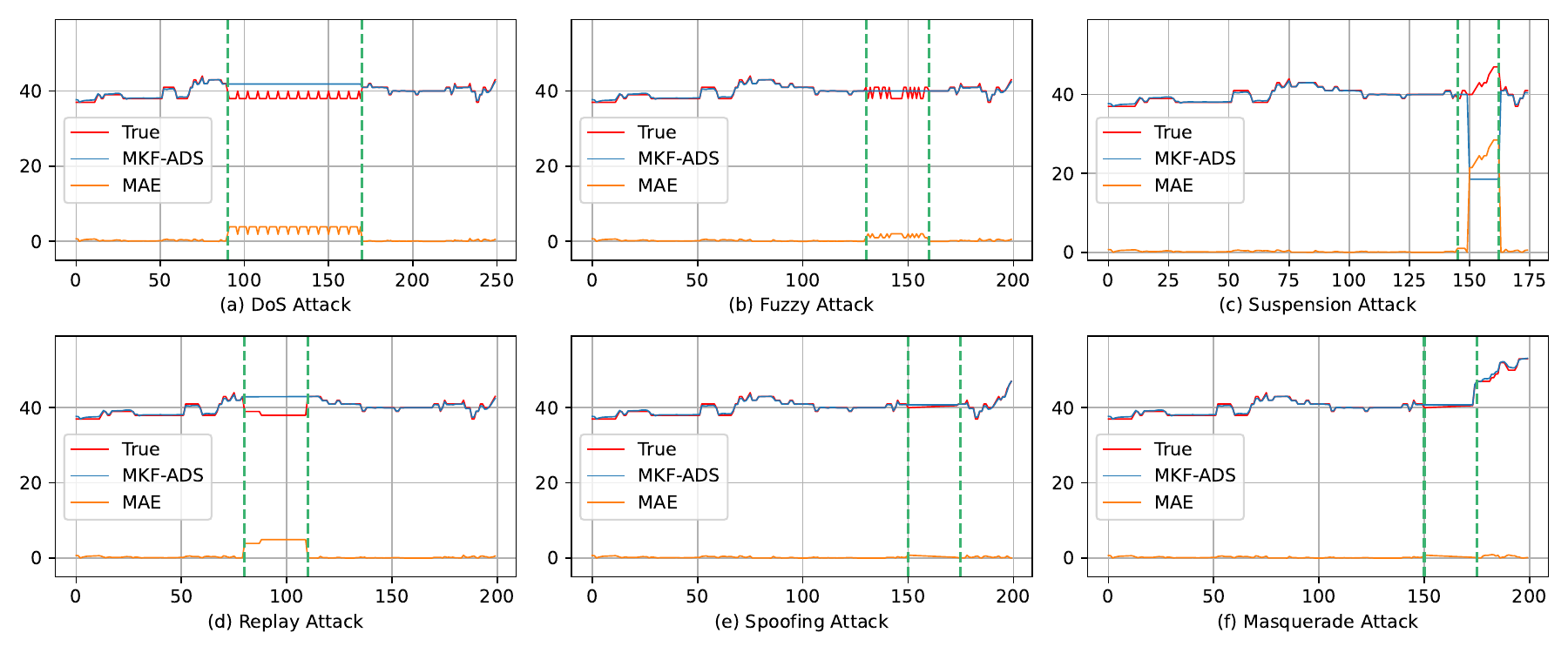}
    \caption{Visualization of the detection performance of fine-grained attack, i.e., the attacker injects the abnormal deviations that are similar to the normal value. The red line is the ground truth of the CAN sequence on the CAN bus, the blue line is the predicted value, and the yellow line is the anomaly score, calculated by MAE.}
    \label{fig:9}
\end{figure*}

After acquiring the predictive capability of MKF-ADS, we set the abnormal threshold to 0.04 to detect as many malicious messages as possible. Next,  we refer to the anomaly sequence generation method in the~\cite{hanselmann2020canet} to inject the attack sequence to the normal CAN sequence randomly in six attack scenarios, where the masquerade attack scenario is complementary. Also,  we inject anomalous sequences by different degrees of deviations from the normal value, in order to analyze the proposed model capability for anomaly identification in terms of coarse and fine-grained. Fig.~\ref{fig:8} and Fig.~\ref{fig:9} present the anomaly detection performance of the model in the coarse-grained and fine-grained, respectively. MKF-ADS can accurately detect anomalous messages when the attacker attempts a coarse-grained attack to significantly alter the state of the vehicle in a short period. On the one hand, the predictions of the model are highly consistent with the normal signals during ECU communication. On the other hand, the predicted and true values show significant deviations during the time interval of injected messages. In this case, MKF-ADS alerts the CAN bus and discards the abnormal message. Specifically, Fig.~\ref{fig:8} (a) shows the detection results of the DoS attack. Due to the injection of a large number of malicious messages in a short period, MKF-ADS successfully identifies this anomaly that frequently exceeds the threshold. In terms of fuzzy attacks, the attacker misleads the bus by injecting random signals. However, once the range of random signals is not controlled, MKF-ADS will detect the anomaly. For suspension and replay attacks, compliant messages always appear on the CAN bus at incorrect times. However, the model can capture this disruption of the timing relationship because the anomalous deviations are from the predicted signals based on the history window, and attack messages rather than the normal messages. In Fig.~\ref{fig:8} (e), the attacker modifies the signal values with a certain purpose to spoof the bus, e.g., modifying the gear speed. In contrast, the forgery attack is more stealthy in that it removes normal messages before inserting malicious ones. However, such anomalies are just as easily checked under MKF-ADS monitoring. 

Subsequently, we measure the model's ability to perform fine-grained analysis by simulating injections of more insidious attacks, i.e., conditions under which the attacker will control the generation of malicious messages. For DoS attacks, we inject numerous messages within normal message boundaries. Although the fluctuation values drop dramatically, the injection pattern of a large number of messages draws the model's attention. In the Fuzzy attack, the attacker generates random data with well-controlled bounds, where the anomaly score ranges from a high of 2.0 to a low of only 0.9. In the suspension attack, the attacker chooses to deactivate the target ECU out of a stable signal. Similarly, the replay attack records a segment of telegrams close to a normal signal and injects it later. Although the anomaly scores are scaled significantly, MKF-ADS still recognizes anomalous intervals. In contrast, we create more insidious spoofing and masquerade attacks. In Fig.~\ref{fig:9}, the elaborately modified signals will be consistent with the normal signals for several numbers, with fluctuations ranging from only 0.77 to 0.19.
We find that our model has only one false alarm on the attack interval in this scenario. Hence, MKF-ADS also maintains stable performance, even if in the fine-grained attack scenario.

\begin{table}[t]
    \centering
    \caption{COMPARISON WITH BASELINE IN SIX ATTACK SCENARIOS}
    \label{tab3}
    \large
    \resizebox{\linewidth}{!}{
    \begin{tabular}{llccccc}
        \toprule
         Attack & Model & ER (\%) & FAR (\%) & R (\%) & P (\%) & F1 (\%)  \\ \midrule
         \multirow{5}{*}{DoS} & LSTM-P~\cite{taylor2016anomaly} & 7.25 & 9.50  & 90.9 & 95.0 & 92.9 \\
         & LSTM-E~\cite{qin2021application} & 5.85  & 7.00 & 93.1 & 95.3 & 94.2 \\
         & DeepConvGRU~\cite{zhang2019deep} & 3.35 & 4.00& 96.0 & 97.3 & 96.6\\
         & CLAM~\cite{sun2021anomaly} & 2.45  & 2.20 & 97.7 & 97.3 & 97.5\\ \cmidrule{2-7}
         & MKF-ADS (Ours) &  \textbf{0.85} & \textbf{0.80} & \textbf{99.1} & \textbf{99.1} & \textbf{99.1}\\ \midrule
         \multirow{5}{*}{Fuzzy} & LSTM-P~\cite{taylor2016anomaly} & 7.85 & 7.80 & 92.1 & 92.1 & 92.1\\
         & LSTM-E~\cite{qin2021application} & 5.87 & 6.80 & 93.2 & 95.0 & 94.1\\
         & DeepConvGRU~\cite{zhang2019deep} &  5.97 & 4.84 & 95.0  &92.9 & 93.9\\
         & CLAM~\cite{sun2021anomaly} & 4.32 &4.84 & 96.2 & 95.2& 95.6  \\ \cmidrule{2-7}
         & MKF-ADS (Ours) & \textbf{1.55} & \textbf{1.30} & \textbf{98.6}& \textbf{98.2} & \textbf{98.4}\\ \midrule
         \multirow{5}{*}{Replay} & LSTM-P~\cite{taylor2016anomaly} & 5.55 &2.30 & 97.5 & 91.2 & 94.3 \\
         & LSTM-E~\cite{qin2021application} & 5.42 & 5.05 & 94.9 & 94.2 & 94.5\\
         & DeepConvGRU~\cite{zhang2019deep} & 4.55 &3.10 &  96.8 & 94.0 & 95.3  \\
         & CLAM~\cite{sun2021anomaly} & 4.67 &5.05 & 94.9 & 95.8 & 95.3\\ \cmidrule{2-7}
         & MKF-ADS (Ours) & \textbf{2.00} & \textbf{1.80} & \textbf{98.1} & \textbf{97.8} & \textbf{97.9} \\ \midrule
         \multirow{5}{*}{Suspension} & LSTM-P~\cite{taylor2016anomaly} & 2.65 & 3.60&  96.4 & 98.3 & 97.3 \\
         & LSTM-E~\cite{qin2021application} & 2.40& 3.60 & 96.4 &98.8 & 97.6 \\
         & DeepConvGRU~\cite{zhang2019deep} &  2.87 &3.70 & 96.3 &97.9 & 97.1 \\
         & CLAM~\cite{sun2021anomaly} &  2.10 & \textbf{1.70}  & 98.2 & 97.5 & 97.8\\ \cmidrule{2-7}
         & MKF-ADS (Ours) & \textbf{1.60} & \textbf{1.70} & \textbf{98.3} & \textbf{98.5} & \textbf{98.4}\\ \midrule 
         \multirow{5}{*}{Spoofing} & LSTM-P~\cite{taylor2016anomaly} & 8.75 &9.20 & 90.8 & 91.7 & 91.3\\
         & LSTM-E~\cite{qin2021application} & 8.25 & 9.20 & 90.9 & 92.7 & 91.8\\
         & DeepConvGRU~\cite{zhang2019deep} & 7.75  &8.20 & 91.8 & 92.7 & 92.3\\
         & CLAM~\cite{sun2021anomaly} & 6.10 & \textbf{5.20} &\textbf{94.7} & 93.0 & 93.8\\ \cmidrule{2-7}
         & MKF-ADS (Ours) & \textbf{4.60}& 5.40 & 94.6 & \textbf{96.2} & \textbf{95.4}\\ \midrule
         \multirow{5}{*}{Masquerade} & LSTM-P~\cite{taylor2016anomaly} & 10.10 & 9.40 & 90.4 & 89.2 & 89.8 \\
         & LSTM-E~\cite{qin2021application} & 9.00 & 9.20 & 90.8 & 91.2 & 91.0 \\
         & DeepConvGRU~\cite{zhang2019deep} & 8.20 & 8.30 & 91.7 & 91.9 & 91.8\\
         & CLAM~\cite{sun2021anomaly} & 6.75 & 6.30 & 93.6 & 92.8 & 93.2\\ \cmidrule{2-7}
         & MKF-ADS (Ours) & \textbf{5.15} & \textbf{3.40} & \textbf{96.4} & \textbf{93.1} & \textbf{94.7} \\ \midrule
    \end{tabular}}
\end{table}
Based on the six attack scenarios set, we validate the model at scale from the detection perspective. Table~\ref{tab3} presents the detection performance compared to the baseline. We find that the MKF-ADS model performs optimally on all metrics with the lowest error rate, FAR, and the highest recall, precision, and F1 scores. However, LSTM-P and LSTM-E considering single temporal knowledge perform the worst, with the highest false alarm rates on various attack scenarios. With the introduction of spatio-temporal knowledge, the DeepConvGRU model has the ability to detect more anomalies, but with poorer robustness.The error of CLAM is the most stable in the baseline, attributed to the attention mechanism focusing on the computation of important time steps. Specifically, the proposed model performs best on DoS attacks with 99.1\% precision and 99.1\% recall, attributed to injecting a large number of malicious messages that cannot escape detection by the model. Also, MKF-ADS maintains 98.2\% precision and 98.6\% recall on fuzzy attacks, attributed to the fact that the random signals are completely distinct from the normal ones. In replay and pause attacks, the injected legitimate messages fail to mislead the context, with precision and recall of 98.3\% vs. 98.3\% and 97.8\% vs. 98.5\%, respectively. We observe a significant performance degradation of MKF-ADS on spoofing detection, owing to the fact that gradual load changes are severely misleading to the context, especially for further masquerade attacks. However, the proposed model can remain highly competitive, achieving a precision and recall of 94.6\% vs. 96.4\% and 96.2\% vs. 93.1\%. Due to the excellent precision and recall, the model achieved F1 scores of 99.14\%, 98.4\%, 97.9\%, 98.4\%, 95.4\%, and 94.7\%, respectively. Correspondingly, the error rate reflects the overall detection ability of MKF-ADS, fluctuating only between 0.85 and 5.15. Moreover, the false alarm rate requires the model to maintain a low level of false positives to maintain vehicle status, thus MKF-ADS reduces it to 0.8-5.4.
\subsection{Generalization performance}
\begin{figure}[t]
    \centering
    \includegraphics[width=1\linewidth]{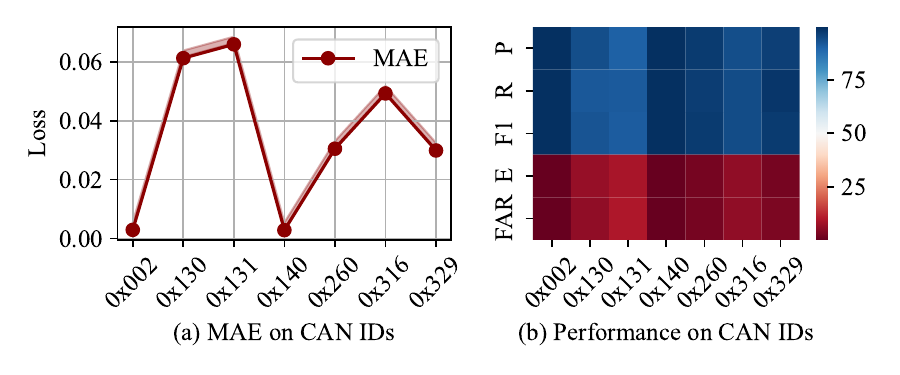}
    \caption{Evaluation on all candidate CAN IDs, where (a) represents the test error, and (b) is the average performance on six attack scenarios.}
    \label{fig:11}
\end{figure}
Further, we evaluate the proposed MKF-ADS on all candidate CAN IDs. Specifically, we obtain the training data through the observation in Fig.~\ref{fig:id}. Thereafter, we train MKF-ADS on these datasets. Fig.~\ref{fig:11} shows the prediction and detection performance. We find that the MAE errors for different CAN IDs are significant, attributed to the degree of fluctuation in the data. For instance, there is a periodic pattern for 0x002 and 0x004, where the error of the model is almost 0. The error for messages similar to 0x260 is controlled to be around 0.03. However, 0x130 and 0x131 have the worst fit, which only reduces to about 0.06. In terms of detection performance, the precision, recall, and F1 scores of all CAN IDs converge to the blue color, i.e. with attack detection capability. Besides, the errors all converge to red, indicating that the FAR and ER are within the controllable range.
\begin{figure}[t]
    \centering
    \includegraphics[width=1\linewidth]{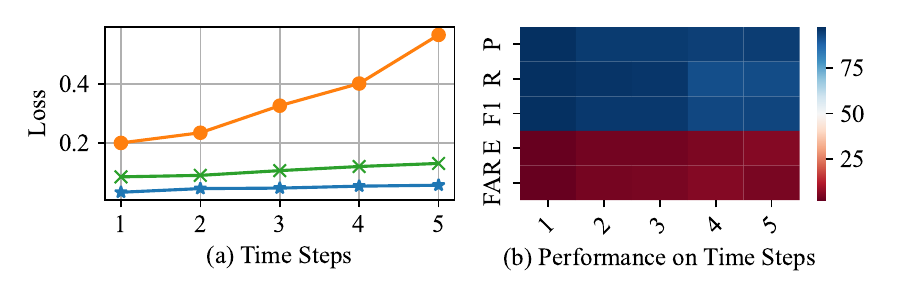}
    \caption{Evaluation on time steps from 1 to 5, where (a) represents the test error, and (b) is the average performance on six attack scenarios.}
    \label{fig:12}
\end{figure}

Given a multivariate time series $X^{(i)}\in \mathbb{R}^{M\times T}$, we can set different predict targets $\hat{y}^{(i)} \in \mathbb{R}^{M\times L}$. Hence, we evaluate the $L$ from 1 to 5, as shown in Fig.~\ref{fig:12}. We find that the prediction error gradually increases as the number of prediction targets increases. For example, when predicting 5 time steps, the MAE reached 0.054. From the detection perspective, we set the prediction to be an attack message once there is a deviation from the threshold in the prediction interval. Similarly, the detection performance saturates while the false alarm rate stabilizes at a low level, albeit with a gradual degradation as the time step increases.

\begin{figure}[t]
    \centering
    \includegraphics[width=1\linewidth]{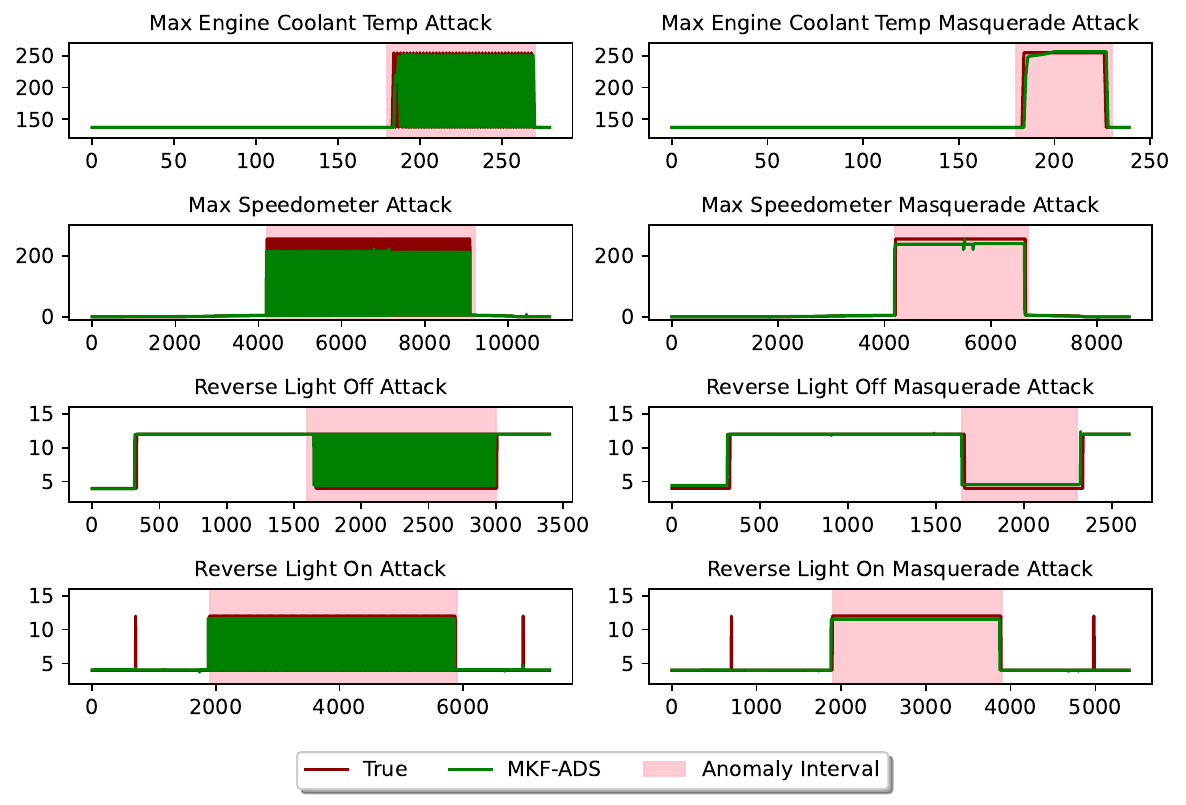}
    \caption{Evaluation of the ROAD dataset, including four spoofing attacks
and four masquerade attacks (related to the Engine, Speedometer, and Light), where the red line is the ground truth, and the green line is prediction of MKF-ADS.}
    \label{fig:13}
\end{figure}

Moreover, we present the generalization performance on the ROAD dataset. In this task, numerous attacks are targeted towards spoofing and masquerade scenarios, which can seriously impact the physical state of vehicles. These effects include the accelerator failing, dashboard lights and headlights turning on, and the seat position shifting. For example, a spoof on 0x0D0 modifies the third bytes to `0C' to change the status of the reverse light from off to on. Fig.~\ref{fig:13} presents the prediction results on spoofing attack and masquerade attack scenarios, related to the Engine, Speedometer, and Light. Based on our observation, the current state of the vehicle is always maintained on the CAN bus for a while. However, when an attacker injects messages that change the behavior of the vehicle (e.g., speed changed to `FF'), MKF-ADS will deviate significantly from the anomalous message during the attack interval for reasons of only knowing the current signal. In short, the proposed model can identify attack periods based on specific thresholds and then alert the bus-layer.

\subsection{Vehicle-Level Model Efficiency Evaluation}
In this subsection, we appraise and compare the application of the model to vehicular networks. Table~\ref{tab5} shows a comparison of the proposed model and baseline in terms of complexity. We report the number of parameters, Floating Point Operations per Second (FLOPs), Multiple Adds (MACs), and memory cost through thop and torchinfo library. Clearly, MKF-ADS achieves fusion modeling of knowledge with only 1748 parametric quantities. In contrast, the number of parameters for the baseline model increases exponentially with the kind of knowledge considered. On FLOPs and MACs, MKF-ADS requires more computational steps. It is a demonstrable fact that the more knowledge is considered, the larger the memory footprint. Thus, the memory footprint of MKF-ADS is 4.05, which is significantly lower than CLAM due to parameter pruning, while higher than the residual baseline. The complexity of the proposed model is within manageable limits for detection performance reasons.
\begin{table}[t]
    \centering
    \caption{Computation cost of the model compared with baseline.}
    \label{tab5}
    \Large
    \resizebox{\linewidth}{!}{
    \begin{tabular}{cccccc}
    \toprule
    Model & \makecell[c]{\#Number \\of Parameters} & FLOPs (M) & MACs (M) & Memory (MB) \\ \midrule
      LSTM-P     &  6,114 & $\boldsymbol{6.2\times10^{-3}}$ & $\boldsymbol{6.1\times10^{-3}}$ & 0.02\\
         LSTM-E & 3,208 & $1.9\times10^{-2}$ & $9.4\times10^{-3}$ & 0.01\\
          DeepConvGRU & 9,304 & $1.8\times10^{-1}$ & $8.9\times10^{-2}$ & 1.16\\
          CLAM & 22,308 & $6.9\times10^{-1}$ & $3.4\times10^{-1}$ & 6.33\\ \midrule
          MKF-ADS (Ours) &\textbf{1,748} &$1.9\times10^{-2}$  &$0.9\times10^{-2}$ & 4.05\\ \bottomrule
    \end{tabular}}
\end{table}
\begin{figure}[t]
    \centering
    \includegraphics[width=1\linewidth]{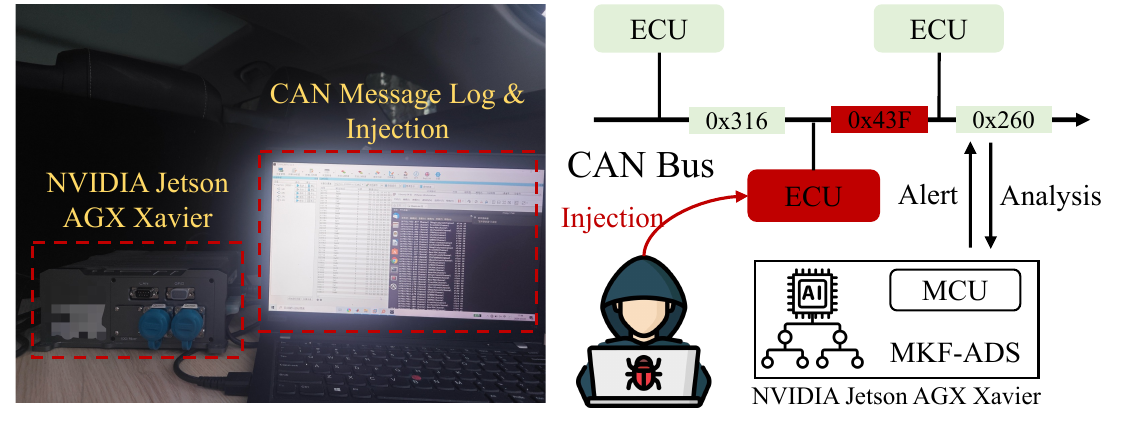}
    \caption{Vehicle-level model efficiency evaluation setting, including the software used for CAN message dump and injection, the hardware used for model inference, and connection between components.}
    \label{fig:10}
\end{figure}

To compute the inference delay, we also set up a vehicle-level efficiency evaluation in a real vehicle (SAIC Roewe Marvel X), as shown in Fig.~\ref{fig:10}. Specifically, we record and inject the CAN messages through the CAN Test software in the PC, which is connected to the OBD-II port via the serial peripheral interface (SPI). Then, we use an intelligent controller with Xavier to deploy the MKF-ADS, where Xavier is an embedded GPU. Table.~\ref{tab6} shows the comparison of the proposed model and baseline in terms of time cost. In this table, DCT, DT, and TT represent the Data Collection Time (mainly pre-processing), Detection Time, and Total Time for each intrusion detection. Based on our observations, LSTM-P consumes more at the DCT, attributed to the complexity of preprocessing. In contrast, the preprocessing steps of MKF-ADS and other baselines are more consistent and only consume from 13.92 ms to 23.12 ms. In terms of detection, the proposed model costs an average of only 2.28 ms, which is much better than LSTM-P and decreases based on CLAM. All in all, the average time cost of 16.2 ms is applicable to protect intelligent connected vehicles.

\begin{table}[t]
    \centering
    \caption{Time cost of the model compared with baseline.}
    \label{tab6}
    \resizebox{\linewidth}{!}{
    \begin{tabular}{ccccc}
    \toprule
    Vehicle & Model &  DCT (ms) & DT (ms) & TT (ms) \\ \midrule
     \multirow{5}{*}{\makecell[c]{SAIC Roewe\\ Marvel x}}& LSTM-P & $\geq$ 72.08  & 11.42 & $\geq$ 83.50  \\
         & LSTM-E  & 18.79 & 3.85 & 22.64\\
         & DeepConvGRU  & 23.12 & \textbf{2.15} & 25.27  \\
         & CLAM &  14.20 & 2.70 & 16.90\\ \cmidrule{2-5}
         & MKF-ADS (Ours) & \textbf{13.92} & 2.28 & \textbf{16.20}\\ \bottomrule
    \end{tabular}}
\end{table}
\section{Conclusion and Future Directions}\label{sec6}
In this paper, we propose a multi-knowledge fusion-based self-supervised anomaly detection model, which can efficiently predict CAN messages that deviate from normal behavior. This model is developed to integrate spatial-temporal correlation with an attention mechanism (STcAM) module and patch sparse-transformer module (PatchST), achieving a knowledge-fused expression on the representation level. The STcAM with fine-pruning uses Conv1D to extract spatial features and subsequently utilizes the Bi-LSTM to obtain the temporal features, of which the attention mechanism will focus on the important time steps. Meanwhile, this work exploits PatchST to capture the combined long-time historical features from independent univariate time series. The framework is based on knowledge distillation to STcAM as a student model for learning intrinsic knowledge and cross the ability to mimic PatchST. As only deploying STcAM, the proposed model is adapted to a resource-limited IVN environment. Compared with the baseline in this paradigm, the prediction error and detection capability are improved and robust.

More contribution is needed to achieve effective anomaly detection based on multivariate time series.  Based on this work, we will further consider that knowledge fused is how to drop noise effects. There is a promised direction, namely interpretative analyses to precisely control the fine-grained integration of knowledge. Moreover, we will strive to improve the forecast performance in terms of multi-time steps, which may realize attack alerts in an early phase. To simplify the complexity of the model, we plan on building an integrating ADS for all CAN IDs, which allows for secure interaction between ECUs.

\bibliography{manuscript}
\bibliographystyle{ieeetr}

\begin{IEEEbiography}[{\includegraphics[width=1in,height=1.25in,clip,keepaspectratio]{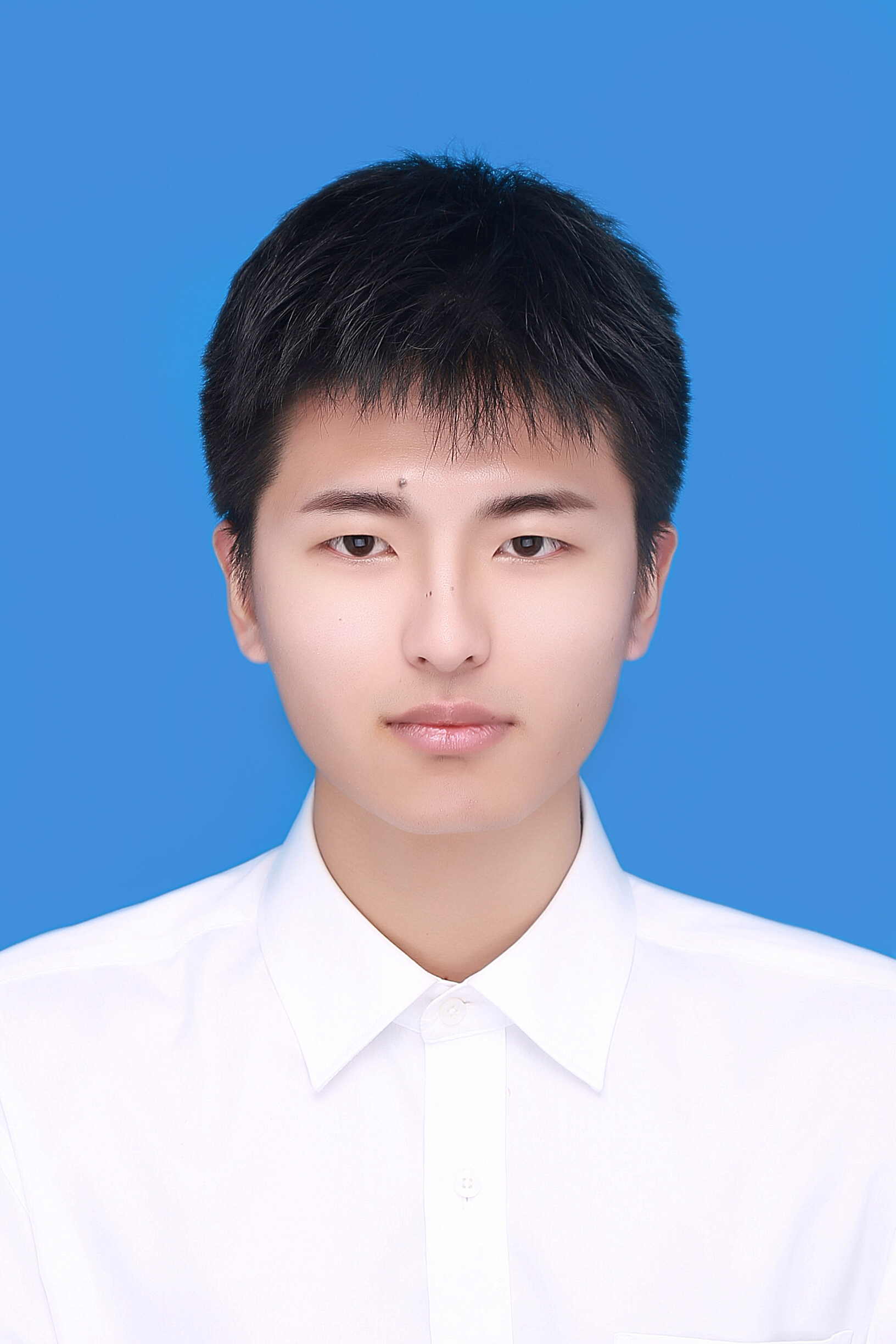}}]{Pengzhou Cheng}
received the M.S. Degree with the Department of Computer Science and Communication Engineering, Jiangsu University, Zhenjiang, China, in 2022. He is currently pursuing the Ph.D. Degree with the Department of Electronic Information and Electrical Engineering, Shanghai Jiao Tong University, Shanghai, 201100, China.

His primary research interests include cybersecurity, machine learning, time-series data analytics, intelligent transportation systems, and intrusion detection system.\end{IEEEbiography}

\begin{IEEEbiography}[{\includegraphics[width=1in,height=1.25in,clip,keepaspectratio]{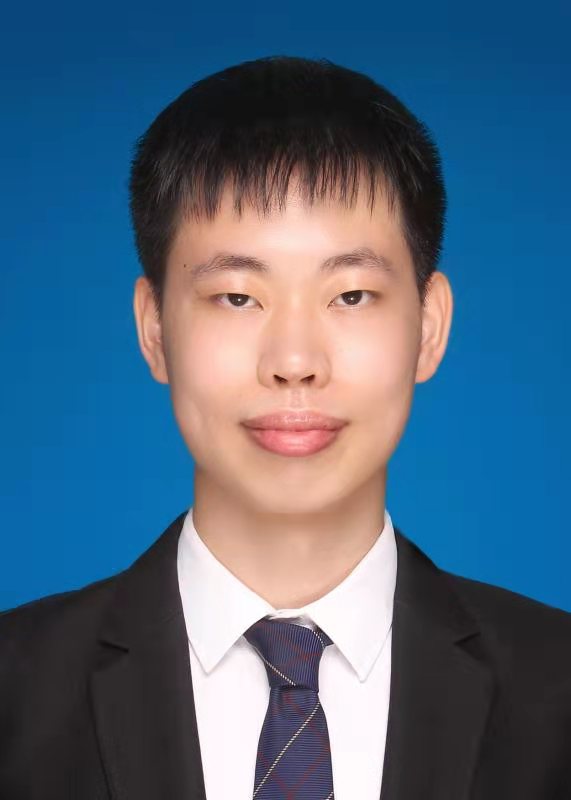}}]{Zongru Wu}
received the B.S Degree from the School of Cyber Science and Engineering, Wuhan University, Hubei, China, in 2022. He is currently persuing the Ph.D. Degree with the School of Cyber Science and Engineering, Shanghai Jiao Tong University, Shanghai, 201100, China.
    
His primary research interests include artificial intelligence security, backdoor attack and countermeasures, cybersecurity, machine learning, and deep learning.
\end{IEEEbiography}

\begin{IEEEbiography}
 [{\includegraphics[width=1in, height=1.25in, clip, keepaspectratio]{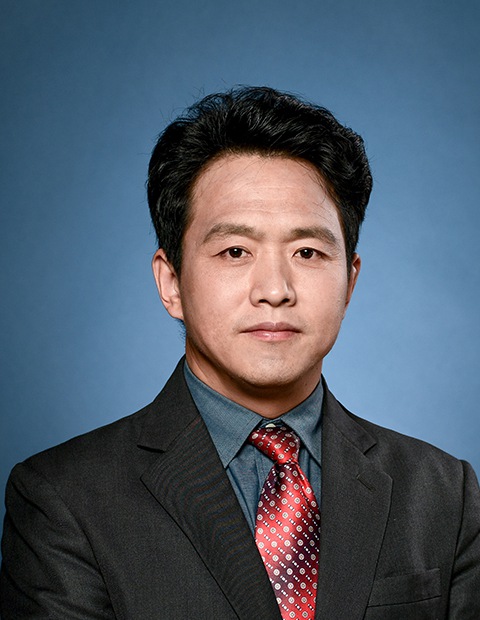}}]{Gongshen Liu}
received his Ph.D. degree in the Department of Computer Science from Shanghai Jiao Tong University. He is currently a professor with the School of Electronic Information and Electrical Engineering, Shanghai Jiao Tong University. His research interests cover natural language processing, machine learning, and artificial intelligence security.
\end{IEEEbiography}

\end{document}